\documentclass[10pt,journal,compsoc]{IEEEtran}

\usepackage{graphicx}
\usepackage[flushleft]{threeparttable}
\usepackage{colortbl}
\usepackage{amsmath}
\usepackage{amsthm}
\usepackage{amssymb}
\usepackage{enumitem}
\usepackage{booktabs}
\usepackage{tabularx}
\usepackage{bm}
\usepackage{bbm}
\usepackage{multirow}
\usepackage{color}
\usepackage[table]{xcolor}
\usepackage{tabu,comment}
\usepackage{array, eucal}
\usepackage[ruled,vlined]{algorithm2e}
\usepackage[american]{babel}
\usepackage{bbding}

\ifCLASSOPTIONcompsoc
  \usepackage[nocompress]{cite}
\else
  \usepackage{cite}
\fi

\ifCLASSOPTIONcompsoc
  \usepackage[caption=false,font=footnotesize,labelfont=sf,textfont=sf]{subfig}
\else
  \usepackage[caption=false,font=footnotesize]{subfig}
\fi

\usepackage[breaklinks=true,colorlinks,citecolor=blue,bookmarks=false]{hyperref}

\definecolor{mygray}{gray}{.9}

\newcommand{\squishlist}{
    \begin{list}{$\bullet$}
        { \setlength{\itemsep}{0pt}
            \setlength{\parsep}{2pt}
            \setlength{\topsep}{2pt}
            \setlength{\partopsep}{0pt}
            \setlength{\leftmargin}{1em}
            \setlength{\labelwidth}{1em}
            \setlength{\labelsep}{0.5em} } }
    \newcommand{\squishend}{
\end{list}  }

\begin{document}

\title{On Point Affiliation in Feature Upsampling}

\author{Wenze~Liu,
Hao~Lu,
Yuliang~Liu,
and~Zhiguo~Cao
\IEEEcompsocitemizethanks{
\IEEEcompsocthanksitem This work is supported by the National Science Foundation of China under Grant No. 62106080.
\IEEEcompsocthanksitem The authors are with School of Artificial Intelligence and Automation, Huazhong University of Science and Technology, Wuhan 430074, China. Corresponding author: H. Lu.\protect\\
E-mail: \{wzliu,hlu,ylliu,zgcao\}@hust.edu.cn
}%
}

%

\IEEEtitleabstractindextext{%

\begin{abstract}

We introduce the notion of point affiliation into feature upsampling. By abstracting a feature map into non-overlapped semantic clusters formed by points of identical semantic meaning, feature upsampling can be viewed as point affiliation---designating a semantic cluster for each upsampled point. In the framework of kernel-based dynamic upsampling, we show that an upsampled point can resort to its low-res decoder neighbors and high-res encoder point to reason the affiliation, conditioned on the mutual similarity between them. We therefore present a generic formulation for generating similarity-aware upsampling kernels and prove that such kernels encourage not only semantic smoothness but also boundary sharpness. This formulation constitutes a novel, lightweight, and universal upsampling solution, Similarity-Aware Point Affiliation (SAPA). We show its working mechanism via our preliminary designs with window-shape kernel. After probing the limitations of the designs on object detection, we reveal additional insights for upsampling, leading to SAPA with the dynamic kernel shape. Extensive experiments demonstrate that SAPA outperforms prior upsamplers and invites consistent performance improvements on a number of dense prediction tasks, including semantic segmentation, object detection, instance segmentation, panoptic segmentation, image matting, and depth estimation. Code is made available at: \url{https://github.com/tiny-smart/sapa}.
\end{abstract}


}

\maketitle

\IEEEdisplaynontitleabstractindextext

\IEEEpeerreviewmaketitle

\IEEEraisesectionheading{\section{Introduction}\label{sec:introduction}}

\IEEEPARstart{F}{eature} upsampling is a fundamental operation used to recover the spatial resolution of features. It is an essential ingredient in many modern dense prediction models~\cite{badrinarayanan2017segnet,lin2017feature,chen18v3,xie2021segformer,cheng2021masked,li2022depthformer}. This operation, however, can eas ily incur ambiguity. For instance, despite our community has reported higher and higher mIoU metrics in semantic segmentation, boundary predictions are mostly unsatisfactory~\cite{kirillov2023segment}. 
We argue that \textit{upsampling matters}. Indeed, as long as a segmentation model uses nearest neighbor (NN) or bilinear interpolation for upsampling, then the two upsamplers would \textit{ipso facto} smooth boundaries. 
An inverse phenomenon can also be observed when applying max unpooling to segmentation where SegNet~\cite{badrinarayanan2017segnet} tends to generate fragile or echinated masks due to the fact that max unpooling introduces massive zero fillings during upsampling. Max unpooling, interestingly, behaves reasonably well in image matting where it helps reconstruct hairs and furs~\cite{xu2017deep}. The sensitivity of different upsamplers in different tasks therefore has inspired a new and interesting research direction: \textit{developing task-agnostic, general-purpose feature upsamplers}. To this end, several dynamic upsamplers have been introduced, including CARAFE~\cite{jiaqi2019carafe,wang2020carafe++}, IndexNet~\cite{lu2019indices,lu2022index}, A2U~\cite{dai2021learning}, and FADE~\cite{lu2022fade}. Yet, we find that they still have certain (types of) task preference. For example, A2U can fall behind bilinear interpolation in segmentation, and IndexNet favors detail-sensitive tasks.

\begin{figure}[t]
	\centering
	\includegraphics[width=\linewidth]{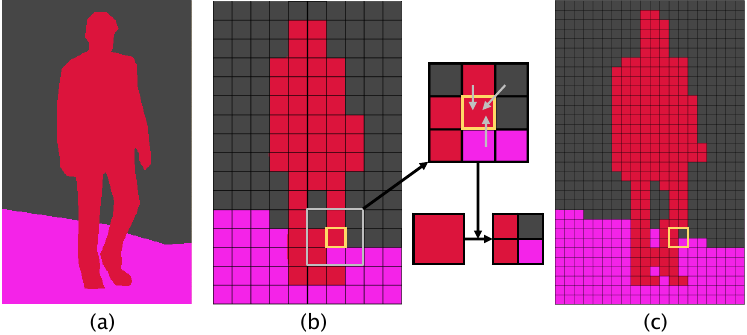}
	\caption{\textbf{Point affiliation.} We consider the low-res (b) and the high-res (c) feature maps are both sampled from an imaginary continuous feature map (a). The feature map is formed by several semantic clusters, \textit{e.g.}, the background (gray), person (red), and road (purple). In each semantic cluster all the points share the same value. We introduce the notion of point affiliation. It depicts the point-cluster correspondence between each upsampled point and its underlying semantic cluster. Considering $\times2$ upsampling, we show that the semantic clusters of the four upsampled points can be inferred with the help of the surrounding semantic clusters in the low-res feature.}
	\label{fig:point_affiliation}
\end{figure}

\begin{figure*}[t]
	\centering
	\includegraphics[width=\linewidth]{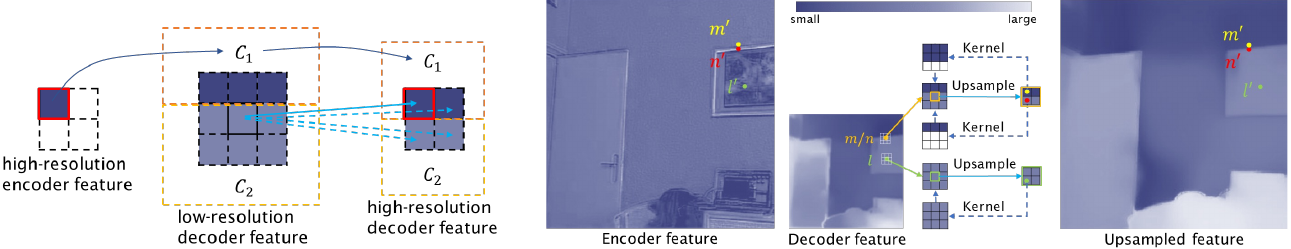}
	\caption{\textbf{Left: Similarity between an encoder point and different semantic clusters in the decoder. Right: Point affiliation mechanism of ideal upsampling kernels.} Left: If the red-box encoder feature point is classified into the semantic cluster $C_1$, then it is more similar to $C_1$ than $C_2$. Right: 
	The upsampling kernels can be a `soft' selector in a local window to assign point affiliation. 
	For an upsampled point, the kernel selects a/some representative points from its most relative semantic cluster. \textit{E.g.}, according to the encoder feature, the red upsampled feature point should belong to the ‘picture' cluster. Then we expect the kernel can assign large weights on picture-related points and small weights on wall-related points. In this way, after the weighted sum, the upsampled point will be revalued and assigned the `picture' cluster.}
	\label{fig:soft_selection}
\end{figure*}

Towards better general-purpose feature upsampling, we introduce the notion of point affiliation. This notion allows us to view and to formulate upsampling from a new perspective. In particular, point affiliation defines a relation between each upsampled point and a \textit{semantic cluster}\footnote{A semantic cluster is formed by local decoder feature points with similar semantic meaning.} to which the point should belong. It highlights the spatial arrangement of upsampled points at the semantic level. Fig.~\ref{fig:point_affiliation} graphically explains this notion. 
From Fig.~\ref{fig:point_affiliation}, we assume that the low-res and high-res feature maps are both sampled from an ideal continuous feature map formed by different clusters. Considering upsampling the red point in Fig.~\ref{fig:point_affiliation}(b) to four upsampled points in Fig.~\ref{fig:point_affiliation}(c), point affiliation is about designating correct semantic clusters to the four points according to available evidence, \textit{e.g.}, the support from surrounding semantic clusters.

\begin{figure}[!t]
	\centering
	\includegraphics[width=\linewidth]{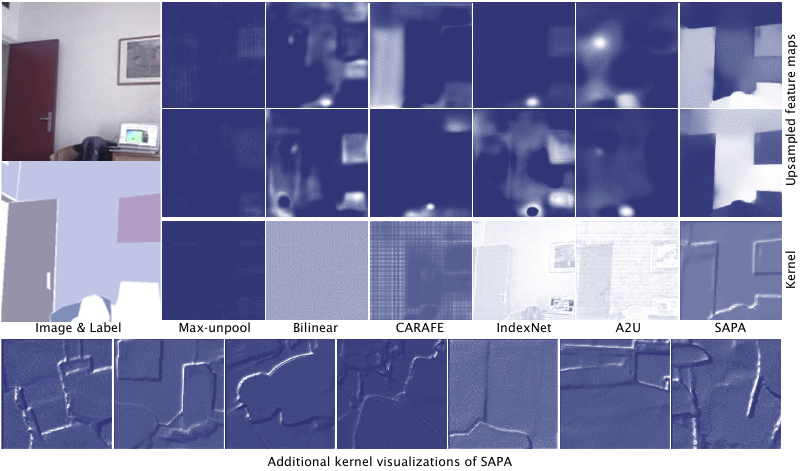}
	\caption{\textbf{Top: Upsampled feature maps and upsampling kernel maps of different upsamplers. Down: Additional upsampling kernel maps of SAPA-B generated from various samples.} The visualization is produced with SegNet~\cite{badrinarayanan2017segnet} as the baseline on the SUN RGBD~\cite{song2015sun} dataset. For each upsampling operator, we choose the first two channels from the feature maps of the last upsampling stage for visualization. Only SAPA-B shows both smooth regions and sharp boundaries. The kernel map of CARAFE is coarse and lacking in details, while IndexNet and A2U generate kernels with much undesired details from the encoder. See the supplementary materials for additional visualizations.}
	\label{fig:upsample_visual}
\end{figure}

Designating point affiliation is difficult and sometimes can be erroneous, however. In $\times2$ upsampling, NN interpolation directly copies four identical points from the low-res one, which assigns the same semantic cluster to the four points. On regions in need of details, the four points probably do not share the same cluster but share anyway. 
Bilinear interpolation assigns point affiliation with distance priors. Yet, when tackling points of different semantic clusters, it not only cannot inform clear point affiliation, but also blurs the boundary between different semantic clusters. Recent dynamic upsamplers have similar issues. CARAFE~\cite{jiaqi2019carafe} judges the affiliation of an upsampled point with content-aware kernels. Certain semantic clusters will receive larger weights than the rest and therefore dominate the affiliation of upsampled points. However, the affiliation near boundaries or on regions full of details can still be ambiguous. As shown in Fig.~\ref{fig:upsample_visual}, the boundaries are unclear in the feature maps upsampled by CARAFE. The reason is that the kernels are conditioned on decoder features alone; the decoder features carry little useful information about high-res structure.

Inferring structure requires high-res encoder features. For instance, if the orange point $m/n$ in Fig.~\ref{fig:soft_selection} lies on the low-res boundary, it is difficult to judge to which cluster the four upsampled points should belong. However, the encoder feature in Fig.~\ref{fig:soft_selection} actually tells that, the yellow point $m'$ belongs to the wall, and the red $n'$ to the picture, which suggests one may extract useful information from the encoder feature to assist point affiliation. Indeed IndexNet~\cite{lu2019indices} and A2U~\cite{dai2021learning} have attempted to encode such information to improve detail delineation in encoder-dependent upsampling kernels; however, the encoder feature can easily introduce noise into kernels, engendering discontinuous feature maps shown in Fig.~\ref{fig:upsample_visual}. Hence, the key problem seems to be how to extract only required information into the upsampling kernels from the encoder feature while filtering out the rest.

To use encoder features effectively, an important assumption of this paper is that, \textit{an encoder feature point is most similar to the semantic cluster into which the point would be 
assigned.} Per the left of Fig.~\ref{fig:soft_selection}, suppose that the encoder point in the red box is assigned into the cluster $C_1$ by its semantic meaning, then it is similar to $C_1$, while not similar to $C_2$. As a result, we can use the mutual similarity score between the encoder feature point and different semantic clusters in the decoder feature as the kernel weights to inform the affiliation of the upsampled point. 
For every encoder feature point, we compute the similarity score between this point and some spatially associated decoder feature points on behalf of the relevant semantic clusters. We simply illustrate the process of generating kernel weights with mutual similarity in case of fixed window-shape kernel, i.e., the spatially associated decoder feature points are selected from a local window. For the green point $l$ in Fig.~\ref{fig:soft_selection}, since every point in the window shares the same semantic cluster, the encoder feature point $l'$ is as similar as every point in the window. In this case we expect an `average kernel' which is the key characteristic to filter noise, and the upsampled four points would have the same semantic cluster as before. For the yellow point $m'$ in the encoder, since it belongs to the `wall' cluster, it is more similar to the points on the wall than those on the picture. In this case we expect a kernel with larger weights on points specific to the `wall' cluster. This can help to assign the affiliation of the yellow point $m'$ to be in the `wall' cluster.

By modeling the local mutual similarity, we derive a generic form of upsampling kernels and show that this form implements our expected upsampling behaviors: encouraging both semantic smoothness and boundary sharpness. Following our formulation, we further instantiate a lightweight, general-purpose upsampler, termed Similarity-Aware Point Affiliation (SAPA). 
The preliminary designs of SAPA, including SAPA-B with embedded inner product similarity and SAPA-I with direct inner product similarity, focus on the generation of kernel weight. With the window-shape kernel 
as in previous dynamic upsamplers~\cite{jiaqi2019carafe,dai2021learning,lu2022fade}, we illustrate the practical upsampling behaviour when using 
the similarity scores as the kernel weights. We also provide a detailed analysis and comparison between two styles of weight generation in SAPA-B and CARAFE at a local level, through the conception of \textit{local continuity and divisibility}. While SAPA-B improves existing upsampling baselines on a number of dense prediction tasks, it is outperformed by CARAFE on object detection. By 
studying the difference between the two tasks, 
we indicate that it is the ability to mend the feature and the large local divisibility in CARAFE that lead to the 
notable performance gain in object detection. To invest such ability 
to SAPA, we move back to the point selection stage. Different from previous upsamplers that select points from a fixed local window, we 
relax it to an arbitrary shape and make it 
feature-dependent. With dynamic point selection, the new variant SAPA-D breaks point affiliation into two steps: 
i) finding 
point candidates, and 
ii) using mutual similarity to 
identify the most representative points. 
Further, 
to trade off between local continuity and divisibility, we 
study the degree of freedom (DOF) in point selection in SAPA-D and customize the 
DOF for different tasks.


We evaluate SAPA across a number of mainstream dense prediction tasks and on both convolutional and transformer architectures, for example: i) \textit{semantic segmentation}: SAPA improves SegFormer-B1~\cite{xie2021segformer} by $2.6$ mIoU and $3.6$ boundary IoU~\cite{cheng2021boundary}, and brings $1.5$ mIoU and $0.9$ boundary IoU improvement for UPerNet-R50~\cite{xiao2018unified};
ii) \textit{object detection}: SAPA boosts $1.8$ AP with Faster R-CNN-R50~\cite{ren2015faster} on MS COCO~\cite{lin2014microsoft}; iii) \textit{instance segmentation}: SAPA improves the performance of Mask R-CNN-R50~\cite{he2017mask} by $1.7$ box AP and $1.3$ mask AP on MS COCO~\cite{lin2014microsoft}; iv) \textit{panoptic segmentation}: SAPA outperforms other upsamplers by large margins and improves $1.9$ PQ with Panoptic FPN-R50~\cite{kirillov2019panopticfpn} on MS COCO~\cite{lin2014microsoft}; v) \textit{image matting}: SAPA outperforms a strong A2U matting baseline~\cite{dai2021learning} on the Adobe Composition-1k testing set~\cite{xu2017deep} with a $4.8\%$ relative error reduction in the SAD metric; vi) \textit{monocular depth estimation}: SAPA reduces the RMS metric of DepthFormer-SwinT~\cite{li2022depthformer} from $0.402$ to $0.392$ on NYU Depth V2~\cite{silberman2012indoor}. Among all variants, the simplest implementations of SAPA outperform or at least are on par with other state-of-the-art dynamic upsamplers, even without additional parameters. Additionally, SAPA-D reports the consistently best performance on the six dense prediction tasks, implying a strong baseline for subsequent upsampling studies. Our contributions are as follows:
\begin{enumerate}
    \item[i)] We introduce the notion of point affiliation for analyzing upsampling behaviors;
    \item[ii)] We propose to designate point affiliation by dynamic kernel shape and local mutual similarity between encoder and decoder features;
    \item[iii)] We provide rich analyses and visualizations of the upsampling behaviors of SAPA and other upsamplers;
    \item[iv)] We present a lightweight and universal upsampler SAPA, investigate its variants,
    and demonstrate its effectiveness and generality across models and tasks. 
\end{enumerate}

A preliminary conference version of this work appeared in~\cite{lu2022sapa}. We extend~\cite{lu2022sapa} from three aspects:
i) we provide formal and precise mathematical definitions for semantic clusters and point affiliation; 
ii) we add detailed 
illustrations, comparisons, and analyses of the upsampling process of SAPA and CARAFE; 
iii) targeting the worse performance than CARAFE on object detection, we provide additional insights and improvements on SAPA by studying the dynamic kernel shape; 
iv) we further extend our experiments on two additional dense prediction tasks, including instance segmentation and panoptic segmentation, to demonstrate the task-agnostic property of SAPA.

\section{Literature Review}
We review work related to dense prediction tasks and feature upsampling.

\subsection{Dense Prediction}
Dense prediction
refers to a wide range of vision tasks that deal with dense per-pixel labelling, from high-level tasks including semantic/instance/panoptic segmentation~\cite{2015Fully,he2017mask,kirillov2019panoptic}, object detection~\cite{girshick2014rich,ren2015faster,2020detr}, and monocular depth estimation~\cite{eigen2014depth,xian2018monocular} to low-level tasks such as image matting~\cite{xu2017deep,lu2019indices,dai2021learning}, image restoration~\cite{dong2015image,zhang2017beyond}, and edge detection~\cite{xie2015holistically}, to name a few. 

The progress of dense prediction is mostly driven by a few representative deep network architectures. For instance, since the introduction of fully convolutional networks (FCNs) in semantic segmentation~\cite{2015Fully}, this field has evolved with many fundamental segmentation architectures such as U-Net~\cite{ronneberger2015u}, SegNet~\cite{badrinarayanan2017segnet}, and DeepLabV3+~\cite{chen18v3}. Similarly in object detection, R-CNN~\cite{girshick2014rich} and YOLO~\cite{Redmon_2016_CVPR} models dominate the main melody in the last few years. Subsequent effect has also been made in multi-scale prediction with typical network architectures such as feature pyramid networks (FPNs)~\cite{lin2017feature}. Many architectures used in other tasks, more or less, have borrowed the designs from segmentation and detection. For instance, the first deep matting model DeepMatting~\cite{xu2017deep} is largely inspired by SegNet; FPNs are also widely used in depth estimation~\cite{xian2018monocular}.

For most representative network architectures above, feature upsampling is an essential ingredient, because most backbone architectures involve downsampling stages, while the expected output is often of high resolution.
An improved, general-purpose upsampler can therefore benefit a significant number of dense prediction models. In this work, we choose a few representative models as baselines in different tasks, including UPerNet~\cite{xiao2018unified} and SegFormer~\cite{xie2021segformer} for semantic segmentation, Faster R-CNN~\cite{ren2015faster} for object detection, Mask R-CNN~\cite{he2017mask} for instance segmentation, Panoptic FPN~\cite{kirillov2019panopticfpn} for panoptic segmentation, A2U Matting~\cite{dai2021learning} for image matting, and DepthFormer~\cite{li2022depthformer} for monocular depth estimation.
We will show that how the notion of point affiliation applies to these tasks and how our upsampler SAPA improves on these baselines.


\subsection{Feature Upsampling}

Standard upsamplers are hand-crafted. NN and bilinear interpolation measure the semantic affiliation in terms of relative distances in upsampling, which follows fixed rules to designate point affiliation, even if the true affiliation may be different. Max unpooling~\cite{badrinarayanan2017segnet} stores the indices of max-pooled feature points in encoder features and uses the sparse indices to guide upsampling. While it executes location-specific point affiliation which benefits detail recovery, most upsampled points are assigned with null affiliation due to zero filling. Pixel Shuffle~\cite{shi2016real} is widely-used in image/video super-resolution. Its upsampling predominantly includes memory operation -- reshaping channel dimension to space dimension. It, to some extent, adjusts point affiliation with prepositioned convolution.

Another stream of upsamplers implement learning-based upsampling. Among them, transposed convolution or deconvolution~\cite{2015Fully} is known as an inverse convolutional operator. Based on a novel interpretation of deconvolution, PixelTCL~\cite{2019Pixel} is proposed to alleviate the checkerboard artifacts~\cite{odena2016deconvolution} of standard deconvolution. In addition, bilinear additive upsampling~\cite{wojna2019devil} attempts to combine learnable convolutional kernels with hand-crafted upsamplers to achieve composited upsampling. Recently, DUpsample~\cite{2020Decoders} seeks to reconstruct the upsampled feature map with pre-learned projection matrices, expecting to achieve a data-dependent upsampling behavior. While these upsamplers are learnable, the upsampling kernels are fixed once learned, still resulting in fixed designation rules of point affiliation.

In learning-based upsampling, some recent work introduces the idea of generating content-aware dynamic kernels. Instead of learning the parameters of the kernels, 
the way to generate the kernels is learned. In particular, CARAFE~\cite{jiaqi2019carafe} predicts dynamic kernels conditioned on the decoder features. IndexNet~\cite{lu2019indices} and A2U~\cite{dai2021learning}, by contrast, generate encoder-dependent kernels. FADE~\cite{lu2022fade} recently unifies the idea of CARAFE and IndexNet and generates upsampling kernels conditioned on both encoder and decoder features. While they significantly outperform previous upsamplers in various tasks, they still can cause uncertain point affiliation, resulting in either unclear predictions near boundaries or fragile predictions in regions.

Our work is closely related to dynamic kernel-based upsampling. We also seek to predict dynamic kernels; however, we aim to address the uncertain point affiliation in prior arts to achieve simultaneous region smoothness and boundary sharpness.

\section{Designating Point Affiliation With Local Mutual Similarity}
In this section we first provide the definition of semantic clusters and point affiliation. Next we revisit the framework of kernel-based dynamic upsampling, where normalized dynamic kernels are typically used to ascertain point affiliation, 
conditioned on local neighborhoods. Then, we 
present our formulation that exploits local mutual similarity to designate point affiliation and showcase our concrete implementations and analyses.

\subsection{Point Affiliation}

We first consider an imaginary continuous feature map $\widetilde{\mathcal{X}}$.
In the sense of sets, $\widetilde{\mathcal{X}}$ 
can be divided into many non-overlapping clusters $\widetilde{\mathcal{S}}_i$, $i=1,2,...,n$. Each cluster $\widetilde{\mathcal{S}}_i$, 
formed by points of the same value $\widetilde{S}_i$, satisfies $\bigcup\limits_i \widetilde{\mathcal{S}}_i=\widetilde{\mathcal{X}}$ and $\widetilde{\mathcal{S}}_i\bigcap \widetilde{\mathcal{S}}_j=\emptyset$ when $i\neq j$. Let $\mathcal{X}$ and $\mathcal{X}'$ be defined as the low-res and high-res discrete feature maps sampled from $\widetilde{\mathcal{X}}$, and the discrete counterparts $\mathcal{S}_i$, $\mathcal{S}_i'$, $S_i$, and $S_i'$ follow the similar definition. We call $\widetilde{\mathcal{S}}_i$, $\mathcal{S}_i$, and $\mathcal{S}_i'$ the semantic clusters, as shown in Fig.~\ref{fig:point_affiliation}. 
The difference between $\mathcal{X}$ and $\mathcal{X}'$ is the distinct density of sampling grids.
We are interested in upsampling $\mathcal{X}$ to $\mathcal{X}'$. In this context, one needs to consider how to determine the cluster value of new appearing points.

According to the concept above, upsampling 
changes $\mathcal{S}_i$ to $\mathcal{S}_i'$, but $\widetilde{\mathcal{S}}_i$ remains unchanged. Thus we formulate upsampling as how to arrange the new appearing points to the correct $\widetilde{\mathcal{S}}_i$, which will then be densely sampled to the new $\mathcal{S}_i'$. We call 
this process in which an upsampled point finds its corresponding semantic cluster as \textit{point affiliation}.

\textit{Remark.} We abstract the notion of semantic cluster from the class label in semantic segmentation. But this concept is universal across different 
dense prediction tasks. In other tasks, the meaning of a semantic cluster may not be an exact class but a more fine-grained semantic part, and the number of clusters can be 
rather large or 
even infinite.

\subsection{Kernel-based Dynamic Upsampling Revisited}

As shown in Fig~\ref{fig:point_affiliation}, beside one single low-res feature point, a local region can include the semantic clusters to which the upsampled four points should belong. Therefore, recent dynamic upsamplers~\cite{jiaqi2019carafe,lu2019indices,dai2021learning} resort to the local neighborhood to designate point affiliation by conducting local dynamic convolution. Their insight is that, for each upsampling point, finding the points of the same semantic cluster in its neighborhood and then generating the upsampled point with a weighted sum. Previous dynamic upsamplers formulate the upsampling process into two components: kernel prediction and feature assembly~\cite{jiaqi2019carafe}, and focus on predicting weights for each window-shape kernel. Here we summarize and extend the process into three steps: point selection, weight generation, and feature assembly.

\vspace{5pt}
\noindent\textbf{Point Selection.} 
When upsampling a certain decoder feature $\mathcal{X}\in\mathbb{R}^{H\times W \times C}$ to the target feature $\mathcal{X}'\in\mathbb{R}^{2H\times 2W \times C}$ (considering an upsampling ratio of $2$), first for each position $l'=(i',j')$ in $\mathcal{X}'$, one selects $S$ points from the neighborhood of $l=(i,j)=\lfloor\frac{l'}{2}\rfloor$ indicating the relevant semantic cluster. Existing dynamic upsamplers including our conference version~\cite{lu2022sapa} set fixed square window kernels to select points of interest, i.e., $I_l=\{l+(u,v):u,v=-r,...,r,r=\lfloor\frac{K}{2}\rfloor\}$, where $K$ is the kernel size. Here we 
relax the window kernel to a more general form: selecting arbitrary points from a neighborhood. Generally, we denote the selected point coordinate and feature point set as $I_{l}=\{\boldsymbol{p}_1, ..., \boldsymbol{p}_S, \boldsymbol{p}_i\in\mathbb{R}^2\}$ and $P_{l}=\{\boldsymbol{x}_1,...,\boldsymbol{x}_S, \boldsymbol{x}\in\mathbb{R}^C\}$,
respectively.

\vspace{5pt}
\noindent\textbf{Weight Generation.}
Then, a kernel map $\mathcal{W}\in\mathbb{R}^{2H\times 2W \times S}$ is generated conditioned on the feature content, which is defined by 
\begin{equation}
    \mathcal{W}={\rm norm}(\psi(\mathcal{Z}))\,,
\end{equation}
where $\mathcal{Z}$ indicates the source feature, $\psi$ refers to a kernel generation module, and ${\rm norm}$ normalizes the $S$ weights at each position. The $\tt softmax$ function is often used as the normalization function such that relevant points can be softly selected to compute the value of the target point. Specifically, in CARAFE $\mathcal{Z}=\mathcal{X}$, while in A2U and IndexNet $\mathcal{Z}=\mathcal{Y}$, where $\mathcal{Y}\in\mathbb{R}^{2H\times 2W \times C'}$ is the high-res encoder feature. According to Fig.~\ref{fig:upsample_visual}, the source for weight generation largely affects the appearance of the predicted kernel maps. With the decoder feature alone, the kernel map is coarse and lacking in details. 
With the encoder feature, the kernel maps generated by IndexNet and A2U have rich details; however, they manifest high similarity to the encoder feature, which means noise is introduced into the kernel. 

\vspace{5pt}
\noindent\textbf{Feature Assembly.}
Each $\boldsymbol{w}'=(w_1, ...,w_S)^T$ at the position $l'$ of $\mathcal{W}$
weighted sums the $S$ points in $P_l$ by 
\begin{equation}
\boldsymbol{x}_{l'}'=\sum\limits_{i=1}^S w_i\boldsymbol{p}_i\,.
\end{equation}
where $\boldsymbol{x}_{l'}'\in\mathbb{R}^{C}$ indicates the upsampled feature point at position $l'$ in $\mathcal{X}'$.
By executing feature assembly on every target position, we can obtain the upsampled feature map. 

Our conference version~\cite{lu2022sapa} focuses on the weight generation stage with a fixed square window shape. As shown in Fig.~\ref{fig:upsample_visual}, the upsampled feature has a close relation to the kernel. 
In our opinion, a well-predicted kernel should encourage not only semantic smoothness but also boundary sharpness; a kernel without encoding details or with too many details encoded can lack detail delineation or introduce noise. We consider an ideal kernel should only response at the position in need of details, while do not response (appearing as an average value over an area) at good semantic regions. 
To this end, for each upsampled point, the corresponding kernel should give larger weights to the correct semantic cluster, while smaller weights to wrong clusters.

In this extension, besides the kernel weights, the kernel shape will be made dynamic, i.e., dynamic point selection. Then the designation of point affiliation would be a two-step strategy: i) (point selection) roughly selecting relevant points for an upsampled position conditioned on the content, and ii) (weight generation) using kernel weights to select the most representative points and weighted sum them to form the upsampled point.

\subsection{Local Mutual Similarity}
\label{subsec:local_mutual_similarity}

We rethink point affiliation from the view of local mutual similarity between encoder and decoder features. With a detailed analysis, we explain why such similarity can assist point affiliation.

We first define a generic similarity function ${\rm sim}(\boldsymbol{x},\boldsymbol{y}): \mathbb{R}^d\times\mathbb{R}^d\rightarrow\mathbb{R}$. It scores the similarity between a vector $\boldsymbol{x}$ and a vector $\boldsymbol{y}$ of the same dimension $d$. We also define a normalization function involving $n$ real numbers $x_1,x_2,...,x_n$ by ${\rm norm}(x_i;x_1,x_2,...,x_n)=\frac{h(x_i)}{\sum_{j=1}^n h(x_j)}$, where $h(x): \mathbb{R}\rightarrow\mathbb{R}$ can be an arbitrary function, ignoring zero division. Given a series of points $\boldsymbol{x}\in P_l$, ${\rm sim}(\boldsymbol{x},\boldsymbol{y})$ and $h(x)$, we can define a generic formulation for generating kernel weights
\begin{equation}
    \label{eq:generic_W}
    w = \frac{h\left({\rm sim}(\boldsymbol{x},\boldsymbol{y})\right)}{\sum\limits_{\boldsymbol{z}\in P_l}h\left({\rm sim}(\boldsymbol{z},\boldsymbol{y})\right)}\,,
\end{equation}
where $w$ is the kernel weight specific to $\boldsymbol{x}$ and $\boldsymbol{y}$. To analyze the upsampling behavior of the kernel, we further define the following notations.

Let $\boldsymbol{y}_{l'}\in \mathbb{R}^{C'}$  denote the encoder feature point at position $l'$ and $\boldsymbol{x}_l\in \mathbb{R}^C$ be the decoder feature point at position $l$, where $C$ is the number of channels. Our operation will be done within between each encoder point $\boldsymbol{y}_{l'}$ and its spatially associated decoder feature point set $P_l$.

To simplify our analysis, we assume \textit{local smoothness}. That is, points with the same semantic cluster will have a similar value. 
Accordingly, a local region 
would share the same value on every channel of the feature map
as well. 
Fig.~\ref{fig:soft_selection} illustrates the case where $P_l$ is selected from a fixed $3\times3$ window. We choose $\boldsymbol{a}\in \mathcal{S}_{wall}$ and $\boldsymbol{b}\in \mathcal{S}_{picture}$ as the feature points 
specific to the `wall' and the `picture' cluster, respectively. Then $\boldsymbol{a}$ and $\boldsymbol{b}$ are both constant points. For ease of analysis, we define two types of windows distinguished by their contents. When all the points inside a window belong to the same semantic cluster, it is called a smooth window; while different semantic clusters appear in a window, it is defined as a detail window.

Next, we explain why the kernel can filter out noise, why it encourages semantic smoothness in a smooth window, and why it can help to recover details when dealing with boundaries/textures in a detail window.

\vspace{5pt}
\noindent\textbf{Upsampling in a Smooth Window.}
Without loss of generality, we consider an encoder point at the position $l'$ in Fig.~\ref{fig:soft_selection}. Its corresponding window is a smooth window of the semantic cluster `picture', thus $\boldsymbol{x}_p=\boldsymbol{b}$, for $\forall p\in I_l=\{l+(u,v):u,v=-r,...,r,r=\lfloor\frac{K}{2}\rfloor\}$. Then the upsampling kernel weight w.r.t.\ the upsampled point $l'$ at the position $p$ takes the form
\begin{equation}
\label{eq:4}
\begin{aligned}
\footnotesize
    w_p&={\rm norm}({\rm sim}(\boldsymbol{x}_p,\boldsymbol{y}_{l'}))\\
    &=\frac{h({\rm sim}(\boldsymbol{x}_p,\boldsymbol{y}_{l'}))}{\sum\limits_{q\in I_l}h({\rm sim}(\boldsymbol{x}_q,\boldsymbol{y}_{l'}))}\\
    &=\frac{h({\rm sim}(\boldsymbol{b},\boldsymbol{y}_{l'}))}{K^2 h({\rm sim}(\boldsymbol{b},\boldsymbol{y}_{l'}))}\\
    &=\frac{1}{K^2}
\end{aligned}\,,
\end{equation}
which has nothing to do with $l$ and $p$. Eq.~\eqref{eq:4} reveals a key characteristic of local mutual similarity in a smooth window: the kernel weight is a constant regardless of $\boldsymbol{y}$. Therefore, the kernel fundamentally can filter out noise from encoder features with an \textit{average} kernel.

In the derivation above, the necessary conditions include: i) $\boldsymbol{x}$ is from a local window in the decoder feature map; ii) a normalization function in the form of $\frac{h(x_i)}{\sum_j h(x_j)}$.

\vspace{5pt}
\noindent\textbf{Upsampling in a Detail Window.}
Again we consider two encoder points at the position $m'$ and $n'$ in Fig.~\ref{fig:soft_selection}. Ideally $\boldsymbol{y}_{m'}$ and $\boldsymbol{y}_{n'}$ should be classified into the semantic cluster of `wall' and 'picture', respectively. Taking $\boldsymbol{y}_{m'}$ as an example, following our assumption, it is more similar to points of the `wall' cluster rather than the `picture' cluster. Then ${\rm sim}(\boldsymbol{x}_s,\boldsymbol{y}_{m'})$ is larger than ${\rm sim}(\boldsymbol{x}_t,\boldsymbol{y}_{m'})$, where $s\in I_a$ and $t\in I_b$, and $I_a$ and $I_b$ are coordinate sets within the `wall' and the `picture' cluster respectively. Here $\{I_a,I_b\}$ is a partition of the window coordinates. Therefore, after computing similarity scores and normalization, one can acquire a kernel with significantly larger weights on points with the semantic cluster of `wall' than that of `picture', \textit{i.e.}, $w_s>>w_t$.
After applying the kernel to the corresponding window, the upsampled point at $m'$ will be revalued and assigned to the semantic cluster of `wall'. Formally, such an assignment process can be interpreted by
\begin{equation}\label{eq:5}
\begin{aligned}
\boldsymbol{x}'_{m'}&=\sum_{s\in I_a}w_s\boldsymbol{x}_s+\sum_{t\in I_b}w_t\boldsymbol{x}_t\\
&=\sum_{s\in I_a}w_s\boldsymbol{a}+\sum_{t\in I_b}w_t\boldsymbol{b}\\
&=\boldsymbol{a}\sum_{s\in I_a}w_s+\boldsymbol{b}(1-\sum_{s\in I_a}w_s)\\
&\to\boldsymbol{a}\,, \text{given that} \sum_{s\in I_a}w_s\to1^-\,.
\end{aligned}
\end{equation}
Similarly, the upsampled point at $n'$ will be assigned into `picture'.
Note that, in Eq.~\eqref{eq:4} we have no constraint on $\boldsymbol{y}$. But in a detail window, $\boldsymbol{y}$ as an encoder feature point can play a vital role for designating correct point affiliation. 

\begin{figure}[!t]
	\centering
	\includegraphics[width=\linewidth]{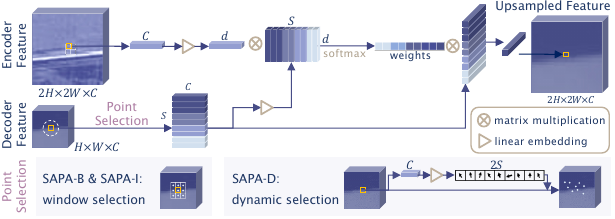}
	\caption{\textbf{Feature upsampling of SAPA.} SAPA follows the process of point selection, weight generation, and feature assembly, from left to right. For an upsampled position, it first selects $S$ relevant point candidates from the spatially associated local decoder feature. Then, it computes mutual similarity scores with inner product between each encoder feature point and the spatially associated local decoder features. The scores are transformed into upsampling kernel weights after kernel normalization. The weights is then used to assemble the selected point candidates. We illustrate two variants: SAPA-B and SAPA-D. They differ in the point selection part. SAPA-B uses the conventional window selection as the previous dynamic upsamplers, and SAPA-D extends it to dynamic selection, where a linear layer generates content-aware point coordinates to select more relevant points in advance. See detailed definition in Section~\ref{subsec:sapa}.
	}
	\label{fig:module}
\end{figure}

\subsection{SAPA-B and SAPA-I: The Preliminary Designs} 
\label{subsec:sapa}

\begin{figure}[!t]
	\centering
	\includegraphics[width=\linewidth]{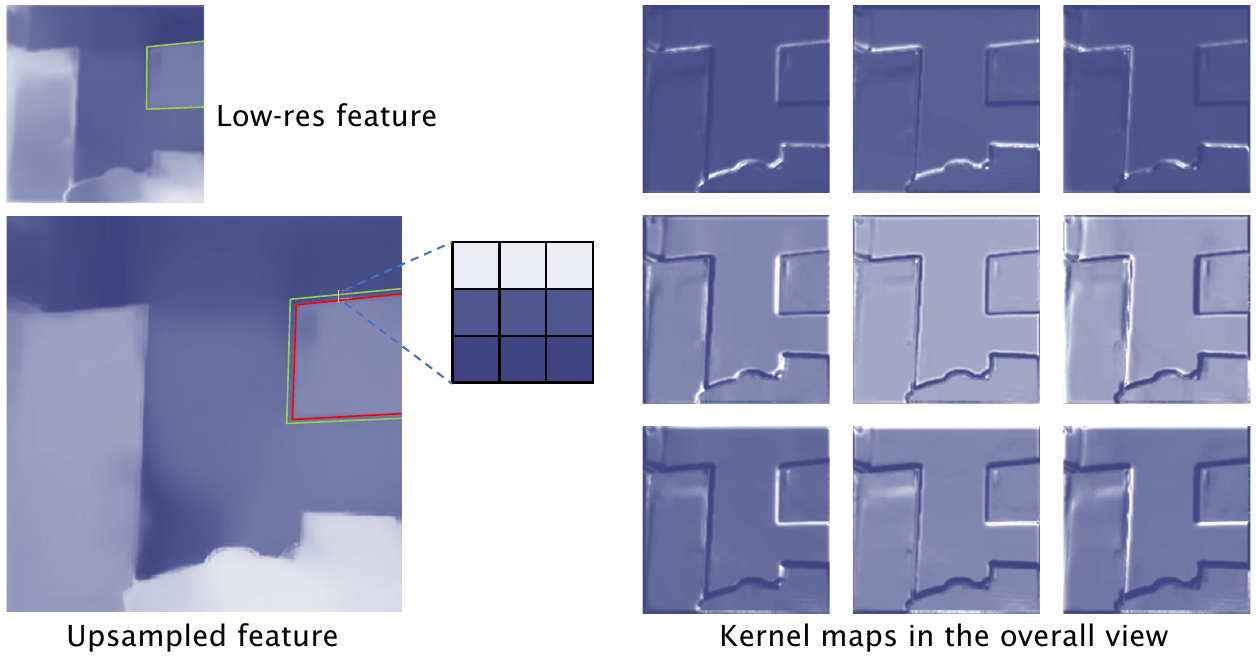}
	\caption{\textbf{The working process of SAPA-B.} We mark the contour of the `picture' in the low-res feature and the upsampled feature with green and red respectively. If one magnify the green-marked low-res feature to the same resolution as the upsmapled feature, then the green box would encircle the red box. Because the contour of the low-res `picture' is blurry and thus fatter than that of the upsampled feature. We consider a certain point between two boxes on the top side. On the right side we visualize each weight in the $3\times3$ kernel predicted by SAPA-B in the overall view. By gathering all values at the same position of the kernel maps, we draw the $3\times3$ kernel in the middle. One can see that the kernel assigns this point to the `wall' cluster.}
	\label{fig:sapa-b_kernels}
\end{figure}

Here we embody the design of SAPA.\footnote{In our conference paper~\cite{lu2022sapa}, we introduce a gating mechanism into the similarity function inspired by~\cite{lu2022fade} to reduce the encoder noise. However, we later find this design only benefits semantic segmentation, while has an adverse effect on instance-level tasks. Therefore, the gating design is deprecated in this journal version.}
The three steps, \textit{i.e.}, point selection, weight generation and feature assembly, of SAPA are shown in Fig.~\ref{fig:module}. We highlight the point selection and kernel generation. After selecting a point set $P_l$ in the decoder feature for each encoder feature point, we compute the similarity score between this point and each point in $P_l$. Similar to self-attention~\cite{vaswani2017attention}, we embody Eq.~\eqref{eq:generic_W} by choosing the inner product similarity and setting the normalization function $h(x)=e^x$, which is equivalent to applying $\tt softmax$ normalization. 

\begin{figure}[!t]
	\centering
	\includegraphics[width=\linewidth]{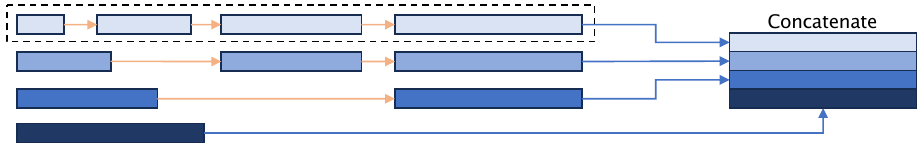}
	\caption{\textbf{The upsampling stages in SegFormer.} The decoder of SegFormer follows an `upsample-then-fuse' manner. In this situation, there are consecutive uspampling stages, e.g., in the dotted box the input feature map will be consecutively upsampled three times. 
 }
	\label{fig:segformer_upsample}
\end{figure}

\begin{figure*}[!t]
	\centering
	\includegraphics[width=\linewidth]{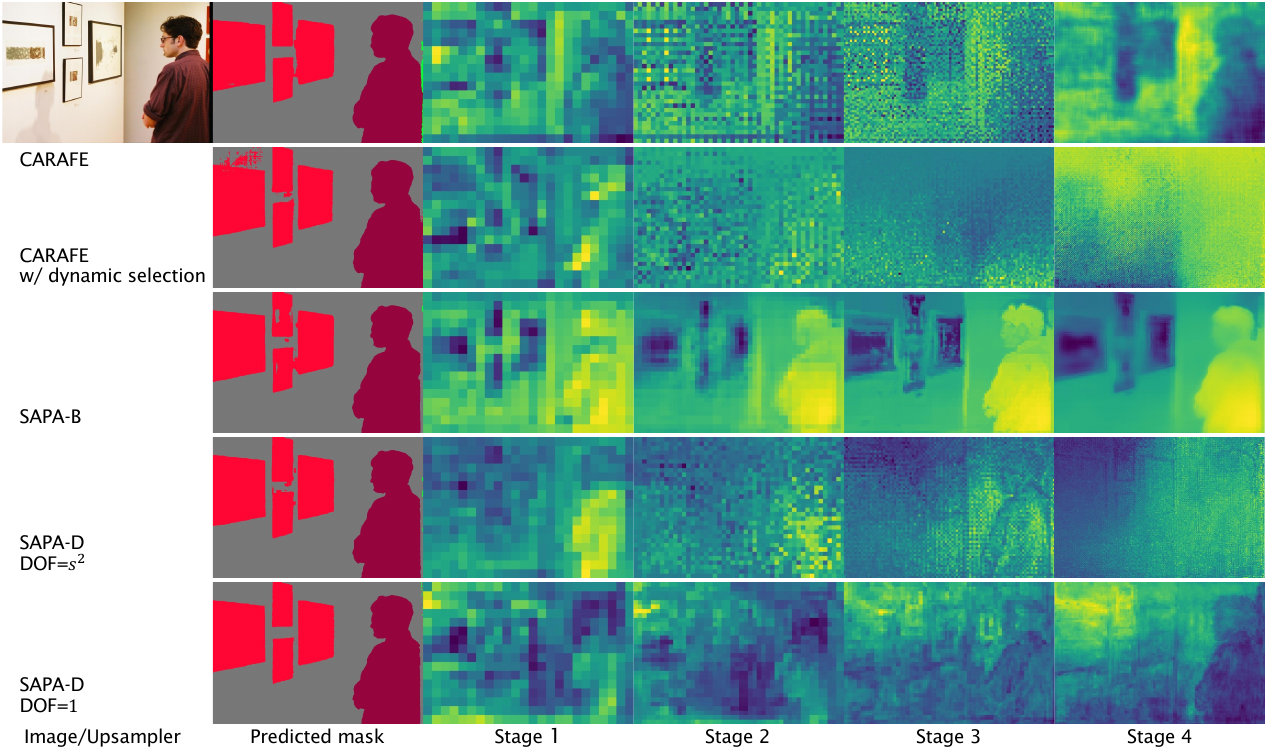}
	\caption{\textbf{Comparison on 
 upsampled feature maps of several upsamplers on semantic segmentation with SegFormer-B1.} 
 For each upsampler, we visualize the predicted mask and the feature maps of four stages as indicated by the dotted box in Fig.~\ref{fig:segformer_upsample}. CARAFE produces continuous interior regions on both the upsampled feature and the predicted mask. However, it introduces local discontinuity on the intermediate features (Stages 3 and 4). In contrast, SAPA-B upsamples the feature map with clear boundaries, but introduces noise from the encoder feature (note that, the noise is defined as the undesired textures and details in the encoder feature; more noise manifests more similarity of the feature map to the encoder feature. After introducing dynamic selection to CARAFE, the upsampled features can even collapse. SAPA-D 
 reduces the noise from the encoder feature (less similar to the encoder feature), but still captures the 
 main structural information. SAPA-D with $\text{DOF}=1$ further eases the local divisibility problem. The feature maps produced by SAPA-D seems rugged, because the dynamic selection strategy influences the position arrangement. Similar to CARAFE, the rugged feature does not affect the predicted mask.
 }
	\label{fig:sapa-b_failure}
\end{figure*}

\begin{figure*}[!t]
	\centering
	\includegraphics[width=\linewidth]{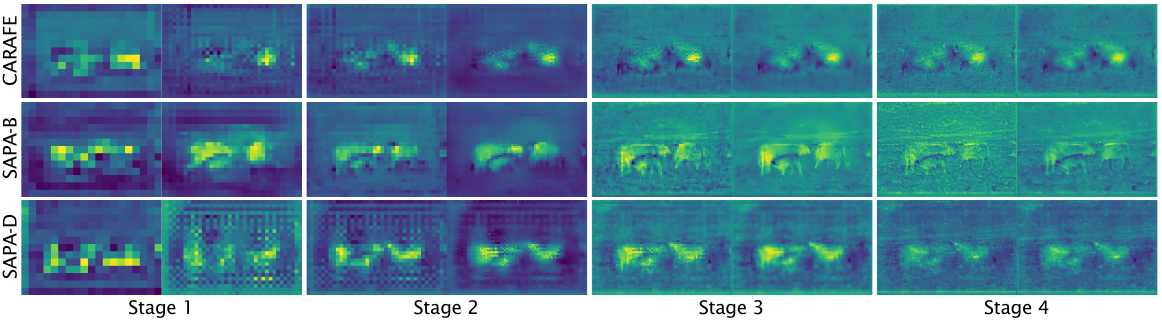}
	\caption{\textbf{Comparison of the input/output features of CARAFE, SAPA-B, and SAPA-D on object detection with Faster R-CNN-R50. } 
    Because of the addition of low-level features in FPN, the semantic clusters in the input feature are disturbed. SAPA-B produces sharp boundaries, while CARAFE highlights some key regions. SAPA-D seems to achieve a compromise between CARAFE and SAPA-B.
 }
\label{fig:faster_rcnn_visual}
\end{figure*}

Based on the discussion of the ideal situation in Section~\ref{subsec:local_mutual_similarity}, we first set $P_l=W_l$ indexed by $I_l=\{l+(u,v):u,v=-r,...,r,r=\lfloor\frac{K}{2}\rfloor\}$ to present a preliminary introduction for the use of mutual similarity to designate point affiliation. Though there are many choices for the similarity function and the normalization function, here we mainly present an empirically well-performed basic design, named SAPA-B (Base), where the encoder and decoder features are linearly embedded to the same dimension $d$, and then the similarity-aware kernel weights are computed with inner product. Mathematically, the upsampled feature point $\boldsymbol{x}'_{l'}$ takes the form
\begin{equation}\label{eq:sapa-b}
\boldsymbol{x}'_{l'}=\sum_{\boldsymbol{x}\in W_l}{\tt softmax}\left(\boldsymbol{x}^T M_{\boldsymbol{x}}^T M_{\boldsymbol{y}}\boldsymbol{y}_{l'}\right)\boldsymbol{x}\,,
\end{equation}
where $M_{\boldsymbol{x}}\in \mathbb{R}^{d\times C}$ and $M_{\boldsymbol{y}}\in \mathbb{R}^{d\times C'}$ are the projection matrices for $\mathcal{X}$ and $\mathcal{Y}$ respectively. In particular, when $C=C'$ we can directly compute the similarity score, 
giving rise to a parameter-free version SAPA-I (Inner product) as
\begin{equation}
\boldsymbol{x}'_{l'}=\sum_{\boldsymbol{x}\in W_l}{\tt softmax}\left(\boldsymbol{x}^T\boldsymbol{y}_{l'}\right)\boldsymbol{x}\,.
\end{equation}

We visualize and compare the upsampling kernel maps and upsampled features in Fig.~\ref{fig:upsample_visual}. Our upsampling kernels show more favorable responses than other upsamplers, with weights highlighted on boundaries and noise suppressed in regions, which visually supports our proposition and is a concrete embodiment of Eq.~\eqref{eq:4} and Eq.~\eqref{eq:5}. 

\subsection{A Closer Look at the Preliminary Designs}

To study the working 
mechanism of SAPA-B, we show how the kernel weights produce one upsampled point in Fig.~\ref{fig:sapa-b_kernels}. We mark the contour of the `picture' in the low-res feature and the upsampled feature in green and red respectively. Because the `picture' in the low-res feature is blurry, the green box is fatter than the red box under the same image size. We choose a certain point on the top side of the `picture' between the green and the red box to study the corresponding kernel weights. On the right side we visualize each weight in the $3\times3$ kernel predicted by SAPA-B in an overall view. By gathering all values at the same position of the kernel maps, we draw the $3\times3$ kernel in the middle. One can see that the kernel assigns this point to the `wall' cluster. The kernel maps also reveals the characteristic of SAPA-B in an overall view, e.g., the top left weight manifests large values on the top left side of an object (or a semantic cluster), while small values on the opposite side in an image.

\vspace{5pt}
\noindent\textbf{
Comparison of Upsampled Feature Maps Between CARAFE and SAPA-B.} We choose an 
example image and study three consecutive upsampling stages in SegFormer 
w.r.t. the dotted box of Fig.~\ref{fig:segformer_upsample}. 
From 
Fig.~\ref{fig:sapa-b_failure}, one can see that CARAFE produces features with continuous interior regions but blurry boundaries, which can also be observed on the predicted mask. On the contrary, SAPA-B predicts sharper boundaries but introduces noise on the `picture' class. More interestingly, a single slice of predicted feature by CARAFE does not have a clear semantic meaning. It implies that different feature slices probably collaborate to work. 
However, a single feature slice generated by SAPA-B already has clear semantic meanings.

\vspace{5pt}
\noindent\textbf{Continuity and Divisibility in Upsampling.}
The feature maps in Fig.~\ref{fig:sapa-b_failure} provides a global understanding of the difference between CARAFE and SAPA-B. Here we study their \textit{local} behaviors. 
For the $s^2$ upsampled neighbors, we 
assume that the trade-off between \textit{continuity} and \textit{divisibility} matters. The continuous property is a rather important prior in natural images, which indicates that 
values usually change continuously over the spatial positions. For an 
operator, we refer to the characteristic of preserving the continuous property of the input feature as `continuity'. 
For instance, 
both the convolution operation in CNNs and the linear projection in MLPs and Transformers maintain the continuous property due to the 
local sharing mechanism. 
That is, the parameters of convolution or projection matrix are shared at each position of the feature/token. On the other hand, the essential goal of an upsampler is `divisibility': 
providing the $s^2$ upsampled neighbors with the possibility to be divided into correct values (semantic clusters) conditioned on the content. 

\begin{figure}[!t]
	\centering
	\includegraphics[width=\linewidth]{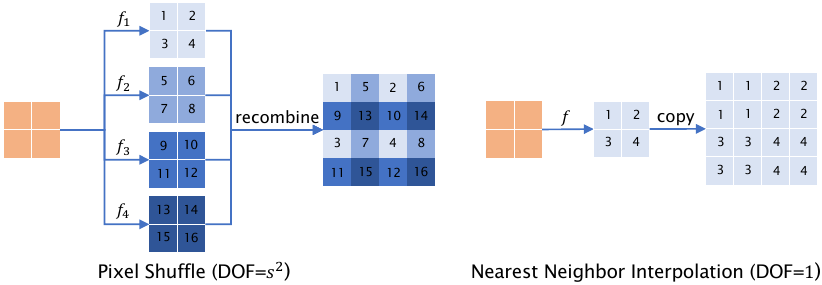}
	\caption{\textbf{The process of Pixel Shuffle and nearest neighbor interpolation.} Different numbers and colors indicate different values and independent generations respectively. In the left figure, four independent functions $f_1$, $f_2$, $f_3$ and $f_4$ generate four feature maps, and Pixel Shuffle recombines them to form the upsampled output. $f_i$'s can be convolutions or linear projections etc. Because of the independence of $f_i$'s, there is local discontinuity in the upsampled feature. We define the DOF of Pixel Shuffle as $s^2$. At the right side, one function $f$ processes the input feature, and NN produces four copies of them to form the upsampled output. We call the DOF of NN as $1$.
 }
	\label{fig:pixelshuffle_process}
\end{figure}

\begin{figure}[!t]
	\centering
	\includegraphics[width=\linewidth]{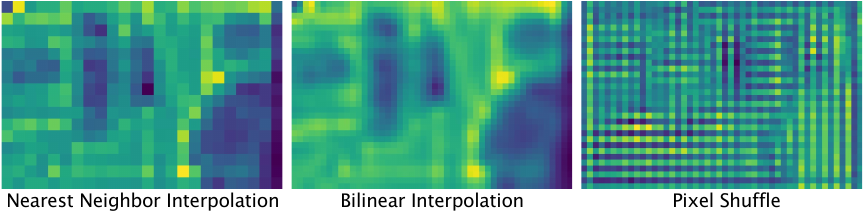}
	\caption{\textbf{Upsampled features produced by interpolation methods and Pixel Shuffle.} Interpolation methods use the distance prior, so they tends to preserve the continuous property in the upsampled feature. But interpolation blurs the edges, 
    resulting in poor divisibility. In contrast, Pixel Shuffle uses independent parameters to generate upsampled neighbors for large divisibility, but it has poor continuity. We consider the DOF of Pixel Shuffle to be four here, because it uses four groups of learnable parameters. 
 }
	\label{fig:bilinear_pixelshuffle}
\end{figure}

There are often conflicts between continuity and divisibility during upsampling, however. For example, because NN or bilinear interpolation uses the distance prior to assign the upsampled values, they produce 
nearly 
continuous feature maps but suffer from weak divisibility 
due to the fixed interpolation rules. In contrast, Pixel Shuffle~\cite{shi2016real} first applies $s^2$ independent convolutions 
to the input and then recombines their outputs to form a feature map $s$ times the size of the input. As shown on the left 
of Fig.~\ref{fig:pixelshuffle_process}, the parameters corresponding to 
certain four upsampled neighbors change independently, so the values of these neighbors have large 
divisibility. These values, however, are prone to 
break the 
continuity. Fig.~\ref{fig:bilinear_pixelshuffle} verifies this point of view by a comparison between two intermediate feature maps produced by nearest neighbor/bilinear interpolation and Pixel Shuffle\footnote{Pixel Shuffle is mainly used in image super-resolution tasks, while hardly seen in other tasks. Since upsampling is often the last step of the network in image super-resolution and the upsampled feature would be directly supervised by the loss, the continuity is guaranteed by the ground truth image. However, upsamplers are usually adopted in the intermediate process in other tasks. Due to the long distance to the supervision, Pixel Shuffle cannot preserve the continuity and tends to cause the checkerboard artifacts.} when processing the same image. 

We now compare CARAFE and SAPA-B when $\times s$ upsampling an input feature in sense of continuity and divisibility. In CARAFE, $s^2$ groups of kernel weights are generated by Pixel Shuffle (Fig.~\ref{fig:pixelshuffle_process}) from the input feature to form $s^2$ upsampling kernels. Because of the independence of the $s^2$ kernels, the local discontinuity appears in the intermediate upsampling stages, as shown in Row 1 of Fig.~\ref{fig:sapa-b_failure}. In contrast, SAPA-B 
creates the high-res kernels by considering the high-res encoder feature, 
where all kernels are generated by the same parameters,\textit{ i.e.}, the projection matrix $M_{\boldsymbol{x}}$ and $M_{\boldsymbol{y}}$. The continuity is ensured by the continuous property of the encoder feature in SAPA-B; it thus produces locally continuous feature maps (Row 3). To sum up, for the transformation 
from low resolution to high resolution, CARAFE uses $s^2$ groups of independent parameters to process the same input for $s^2$ times and recombines them; SAPA-B 
uses the high-res structure of the encoder feature, with only one group of parameters. 
Therefore, the continuity of SAPA-B is better than CARAFE, but the divisibility of CARAFE is better (in SAPA-B the variations of the $s^2$ neighbors are constrained by the encoder feature, while in CARAFE the $s^2$ neighbors can vary arbitrarily via learning). Next we show how we strengthen the divisibility of SAPA-B in another variant.

\subsection{SAPA-D: Dynamic Kernel Shape for SAPA}
\label{subsec:sapa-d}

 \begin{figure*}[!t]
	\centering
	\includegraphics[width=\linewidth]{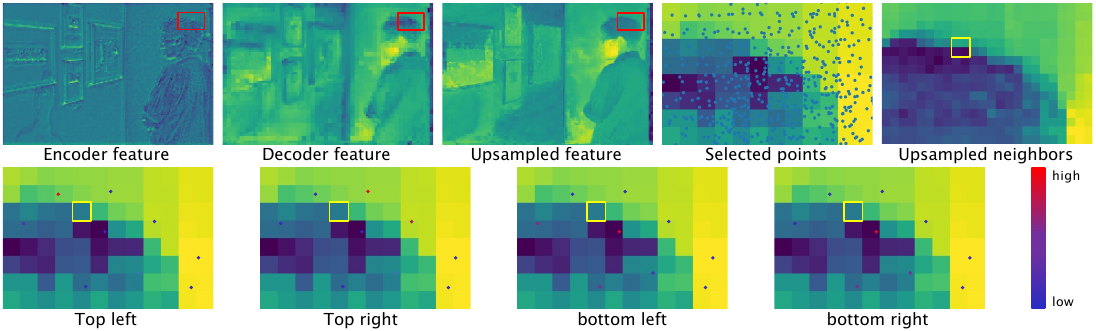}
	\caption{\textbf{Local view of the upsampling process of SAPA-D.} The decoder feature and the upsampled feature correspond to the second and third feature map in Row 2 in Fig.~\ref{fig:segformer_upsample}. 
    A red boxed local region is highlighted to study how the boundary becomes clearer. We mark a point on the boundary of the low-res feature and plot the selected point candidates with small circles. The color of the circle denote the value of the mutual similarity score.}
	\label{fig:sapa-d_process}
\end{figure*}

The preliminary design SAPA-B 
performs worse than CARAFE on object detection~\cite{lu2022sapa}. Here we first point out the uniqueness of object detection and analyze the limitations of SAPA-B compared with CARAFE. Next we provide the solution of dynamic point selection and present the new variant SAPA-D. 

\vspace{5pt}
\noindent\textbf{The Weakness of SAPA-B on Object Detection.} First we 
compare semantic segmentation against object detection. Semantic segmentation is 
a typical dense prediction task where each pixel of the output 
contributes to the evaluation metrics or the losses. However, object detection 
outputs the box location and the object category, 
where 
only several small regions contribute to the losses for a certain object~\cite{carion2020end}. In this sense, 
semantic segmentation and object detection can be considered the pointwise and the non-pointwise task, respectively. With this notion, we summarize the merits of CARAFE on object detection as i) the ability to mend the features and ii) the large local divisibility. We explain them as follows. For the first merit, Section~\ref{subsec:local_mutual_similarity} and Fig.~\ref{fig:sapa-b_kernels} imply that the content-awareness of SAPA-B is dominated by the low-res decoder feature. That is, if the decoder feature has correct semantic clusters like in Fig.~\ref{fig:sapa-b_kernels}, then SAPA-B would boost the performance; but if there already exist artifacts, SAPA-B would introduce more noise. In other words, once the ideal assumption in Section~\ref{subsec:local_mutual_similarity} is broken, SAPA-B would have little benefit. 
As shown in Row 3 of Fig.~\ref{fig:sapa-b_failure}, SAPA-B even 
enhances the artifacts on the `picture'. 
In short, SAPA-B only extracts already existed knowledge from the encoder feature, but cannot learn new knowledge to enhance features. However, the dynamic convolution in CARAFE 
learns new representations based on the decoder feature and is able to mend the features. 
For the second merit, the local divisibility of CARAFE is stronger than that of SAPA-B. Object detection is a non-pointwise task, and only a few feature points play 
roles in detecting objects. In this case 
the whole precise boundary may not boost as much performance as some key interior regions produced by CARAFE, as shown in Fig.~\ref{fig:faster_rcnn_visual}. Apart from the two subjective reasons above, the FPN~\cite{lin2017feature} architecture in Faster R-CNN also 
provides some objective reasons. As one can see in Fig.~\ref{fig:faster_rcnn_visual}, first the FPN has the ability to compensate the high-res details by adding the high-res encoder feature after each upsampling stage, which weakens the requirement of high-res information introduced by SAPA-B. Moreover, due to the addition of low-level encoder features, a number of semantic clusters could be affected by noise, which brings 
additional difficulties for the mutual similarity to decide the correct semantic clusters.

\vspace{5pt}
\noindent\textbf{Dynamic Kernel Shape for SAPA.} Based on the analyses above, 
our idea to improve SAPA is to 
enhance the ability to mend the features and to search for 
potentially correct point candidates in noise. 
We address them in point selection: 
an important yet overlooked stage by all previous dynamic upsamplers. Our solution is to extend the kernel shape from the fixed window to an arbitrary shape, \textit{i.e.}, dynamically selecting relevant points conditioned on the feature content. We use the decoder feature to generate sampling positions, akin to using the decoder feature to learn kernel weights in CARAFE. Technically, the sampling process shares a similar spirit with Deformable Convolutional Networks~\cite{dai2017deformable} (DCNs), where a linear branch projects the decoder feature to generate offsets. We introduce such a dynamic point selection mechanism to SAPA-B, and name the new variant as SAPA-D (Dynamic or Deformable). The process of SAPA-D can be described by
\begin{equation}\label{eq:sapa-d}
\begin{aligned}
I_l&=\phi(\mathcal{X})\,,P_l={\tt bilinear\_sampling}\left(\mathcal{X}, I_l\right)\,,\\
\boldsymbol{x}'_{l'}&=\sum_{\boldsymbol{x}\in P_l}{\tt softmax}\left(\boldsymbol{x}^T M_{\boldsymbol{x}}^T M_{\boldsymbol{y}}\boldsymbol{y}_{l'}\right)\boldsymbol{x}\,,
\end{aligned}
\end{equation}
where $\phi$ is mainly composed of a linear layer, and ${\tt bilinear\_sampling}\left(\mathcal{X},I_l\right)$ denotes bilinearly sampling feature points according to the coordinate set $I_l$ from $\mathcal{X}$. SAPA-D combines the advantages of guided (SAPA-B) and unguided (CARAFE) dynamic upsamplers: the mutual similarity utilizes the structural information of high-res encoder feature; the decoder feature-driven point sampling enables further representation learning. Then, the process of point affiliation is broken into two steps: screening relevant point candidates first, and then exploiting the mutual similarity to select the most representative points. 


\vspace{5pt}
\noindent\textbf{Degree of Freedom of Point Selection in SAPA-D.} 
Albeit imprecise, we 
use the term \textit{Degree of Freedom (DOF)} to somehow reflect the trade-off between local continuity and divisibility. If the $s^2$ values are generated by the same parameters, we define the DOF as $1$. Otherwise, if they are generated with $s^2$ groups of different parameters, the DOF is $s^2$. For example, the DOF of CARAFE is $s^2$, while that of SAPA-B is $1$. The higher DOF means lower continuity and higher divisibility, and vice versa. Since the offsets are generated from the decoder feature, there are two options 
of generation to match the resolution. As shown in Fig.~\ref{fig:pixelshuffle_process}, we can either use Pixel Shuffle or 
NN interpolation to generate the offset for each upsampled position. The offset $\text{DOF}=s^2$ if Pixel Shuffle is used, and $\text{DOF}=1$ if using 
NN interpolation. For example, we 
set the offset $\text{DOF}=1$ for semantic segmentation by default, \textit{i.e.}, each $s^2$ neighbors share the same offsets, while setting $\text{DOF}=s^2$ for instance-level tasks like object detection. Other settings can be referred to Table~\ref{tab:hyper-parameter}. For semantic segmentation with SegFormer-B1, we visualize the predicted masks and the intermediate feature maps produced by SAPA-D with $\text{DOF}=s^2$ and $\text{DOF}=1$ in Row 4 and Row 5, respectively. One can see that large DOF introduces local discontinuity on both the feature maps and the output mask. With $\text{DOF}=1$, SAPA-D 
generates features with good trade-off between continuity and divisibility and outputs a high-quality mask.

\vspace{5pt}
\noindent\textbf{Effect of the High-res Guidance.}
The high-res guiding feature not only provides the details, but also the structural information to stabilize the feature map. As shown in Fig.~\ref{fig:sapa-b_failure} Row 2, we add the dynamic point selection into CARAFE and visualize the output feature maps. 
By comparing Row 1 against Row 2, the absence of high-res guidance causes the collapse of upsampled feature map, and the output mask manifests serious checkerboard effect. Note that, the model still can output an 
approximately correct mask because other features are involved via concatenation (Fig.~\ref{fig:segformer_upsample}).

We visualize the upsampling process of SAPA-D in Fig.~\ref{fig:sapa-d_process}. We highlight a red boxed region to show how the boundary is made clearer by SAPA-D. First, in the red box all selected points are plotted for 
an overall interpretation. Then, we mark a low-res point on the decoder feature with a yellow box, and plot the selected points. We use different colors to represent the mutual similarity scores produced by the encoder feature point and the selected decoder points. From the bottom subfigures, one can see that each upsampled neighbor assigns large weights to the correct semantic cluster.

\subsection{Other Design Details}
Here we provide other minor details of our module design.

\vspace{5pt}
\noindent\textbf{Grouped Upsampling.} Increasing the number of selected points in exchange for performance is 
rather expensive 
in terms of latency. Grouped upsampling~\cite{liu2023learning} is an economic way 
to promote the performance for SAPA-D. We reduce the number of points 
from $25$ in SAPA-B to $9$ in SAPA-D, and apply grouped ($4$ groups) upsampling to 
enhance performance with 
a low latency. Interestingly, the grouping operation is only useful for SAPA-D with dynamic selection, while even hurting the performance for SAPA-B and CARAFE.

\vspace{5pt}
\noindent\textbf{Initialization of Point Selection.} DCN~\cite{dai2017deformable} sets the initial offset as the window shape similar to common convolution. However, we find that setting the initial offset being zeros (called origin initialization) will slightly 
boost the performance. We think it releases the constraint of grid prior and better 
exploits the learning ability.

\vspace{5pt}
\noindent\textbf{Normalization Before Computing Similarity Scores.} When training SegFormer~\cite{xie2021segformer} with SAPA, we observe that the loss begins to increase at around $4000$ iterations. We think this is because in SegFormer certain features need to be consecutively upsampled (Fig.~\ref{fig:segformer_upsample}), the distribution diversity between the encoder and decoder features would have a large effect on the stability of similarity computation. 
For SegFormer, we therefore 
apply {\rm groupnorm}\footnote{In our conference version~\cite{lu2022sapa} we apply {\rm layernorm} on SAPA-B. However it is not so friendly for channel-first tensors on latency. Here we replace it with cheaper {\rm groupnorm}.} before computing similarity scores. Similarly, we observe the gradient vanishing problem when training DepthFormer~\cite{li2022depthformer} (but not always), and we also apply {\rm groupnorm} for DepthFormer.

\subsection{Computational Complexity}

\begin{figure}[!t]
	\centering
	\includegraphics[width=\linewidth]{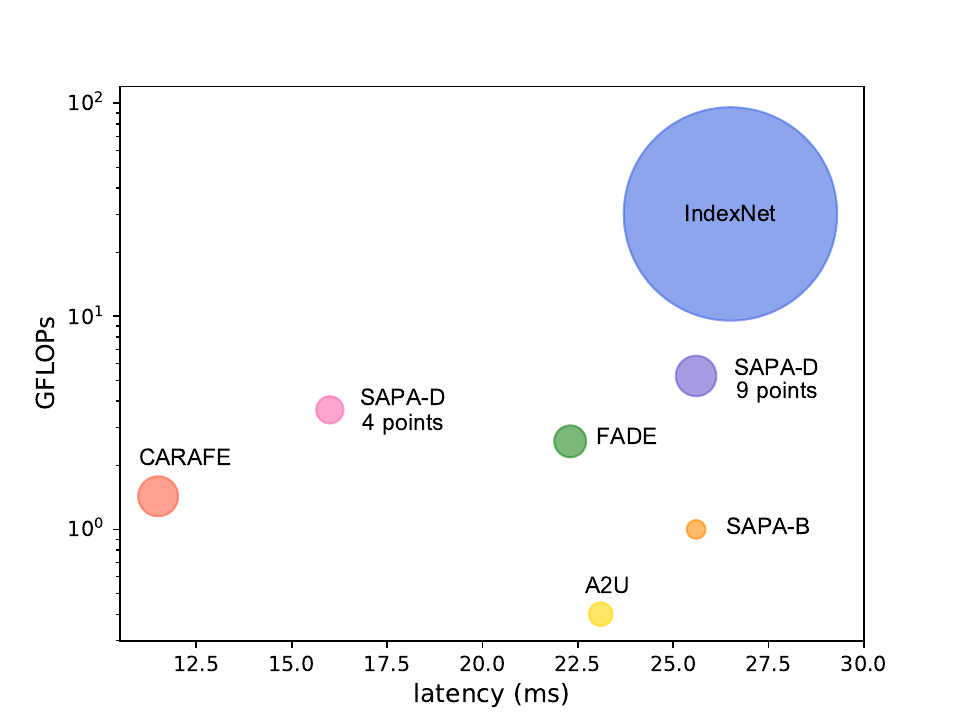}
	\caption{\textbf{Comparison on the practical GFLOPs, number of parameters, and inference time of different dynamic upsamplers.} The circle size indicates the number of parameters. We use the low-res random tensor of size $256\times120\times120$ as input (an auxiliary high-res tensor of size $256\times240\times240$ is provided as required). The inference time is reported on a single NVIDIA GeForce RTX 3090. The 4-point SAPA-D reports similar performance to the 9-point version on Faster R-CNN-R50 (Table~\ref{tab:ablation_point_group}).}
	\label{fig:latency}
\end{figure}

Aside from effectiveness, efficiency is also a concern. Being a part of the overall architecture, an upsampler should not significantly increase the computational overhead. Fortunately, SAPA is lightweight. We calculate the theoretical FLOPs and number of parameters in Table~\ref{tab:complexity}. To showcase its lightweight property, we compare the practical GFLOPs, number of parameters and latency when $\times 2$ upsampling a feature map of size $256\times120\times120$ in Fig.~\ref{fig:latency}. 

\begin{table}[!t]
    \caption{The calculation of FLOPs and parameter numbers for each dynamic upsampler. We name point selection, weight generation and feature assembly as step I, II and III respectively. $C$: channel dimension (for simplification we assume that the encoder feature has the same channel dimension), $d$: embedded dimension, $K$: window kernel size, $S$: selected number of points, $g$: upsmampling groups. $d=64$ and $K=5$ in CARAFE and FADE, while $K=3$ in A2U. We set $d=32$, $S=9$ and $g=4$ in SAPA. 
    }
    \label{tab:complexity}
    \centering
    \renewcommand{\arraystretch}{1}
    \addtolength{\tabcolsep}{-1pt}
    \begin{tabular}{@{}lccc@{}}
    \toprule
        Module & Step & FLOPs ($\times HW$) & Params \\
        \midrule
        CARAFE & II & $Cd+36K^2d$ & $Cd+36K^2d$ \\
         & III & $4K^2C$ & $0$ \\
         & \textbf{Total} & $Cd+36K^2d+4K^2C$ & $Cd+36K^2d$ \\
        \midrule
        IndexNet & II & $32C^2+8C$ & $32C^2+8C$ \\
        HIN & III & $4C$ & $0$ \\
         & \textbf{Total} & $32C^2+12C$ & $32C^2+8C$ \\
        \midrule
        IndexNet & II & $68C^2$ & $68C^2$ \\
        M2O & III & $4C$ & $0$ \\
         & \textbf{Total} & $68C^2+4C$ & $68C^2$ \\
        \midrule
        A2U & II & $73C+4K^2$ & $4K^2C+2C$ \\
         & III & $4K^2C$ & $0$ \\
         & \textbf{Total} & $73C+4K^2+4K^2C$ & $4K^2C+2C$ \\
        \midrule
        FADE & II & $5Cd+45K^2d$ & $2Cd+9K^2d$ \\
         & III & $4K^2C$ & $0$ \\
         & \textbf{Total} & $5Cd+45K^2d+4K^2C$ & $2Cd+9K^2d$ \\
        \midrule
        SAPA-I & II & $4K^2C$ & $0$ \\
         & III & $4K^2C$ & $0$ \\
         & \textbf{Total} & $8K^2C$ & $0$ \\
        \midrule
        SAPA-B & II & $5Cd+4K^2d$ & $2Cd$ \\
         & III & $4K^2C$ & $0$ \\
         & \textbf{Total} & $5Cd+4K^2d+4K^2C$ & $2Cd$ \\
        \midrule
        SAPA-D & I & $32Sdg+32SC+8SCg$ & $8SCg$ \\
         & II & $5Cdg+4Sdg$ & $2Cdg$ \\
         & III & $4SC$ & $0$ \\
         & \textbf{Total} & $5Cdg+38Sdg+36SC+8SCg$ & $2Cdg+8SCg$\\
    \bottomrule
    \end{tabular}
\end{table}

\section{Experiments}
\label{sec:experiments}
\begin{table}[!t]
    \caption{The hyper-parameter settings in SAPA-D for each model.
    }
    \label{tab:hyper-parameter}
    \centering
    \renewcommand{\arraystretch}{1}
    \addtolength{\tabcolsep}{-2pt}
    \begin{tabular}{@{}lcccc@{}}
    \toprule
        Model & \#points & \#groups & offset DOF & groupnorm \\
        \midrule
        SegFormer~\cite{xie2021segformer} & 9 & 4 & 1 & \Checkmark \\
        UPerNet~\cite{zhou2017scene} & 9 & 1 & 1 & \XSolidBrush \\
        Faster R-CNN~\cite{ren2015faster} & 9 & 4 & $s^2$ & \XSolidBrush \\
        Mask R-CNN~\cite{he2017mask} & 9 & 4 & $s^2$ & \XSolidBrush \\
        Panoptic R-CNN~\cite{kirillov2019panopticfpn} & 9 & 4 & $s^2$ & \XSolidBrush \\
        A2U matting~\cite{dai2021learning} & 9 & 1 & 1 & \Checkmark \\
        DepthFormer~\cite{li2022depthformer} & 9 & 4 & $s^2$ & \Checkmark \\
        
    \bottomrule
    \end{tabular}
\end{table}
We first showcase the universality of SAPA across six dense prediction tasks, including semantic segmentation, object detection, instance segmentation, panoptic segmentation, depth estimation, and image matting. For comparing upsamplers, we set the kernel size in Deconvolution and Pixel Shuffle~\cite{shi2016real} as $3$. We use the `HIN' version of IndexNet~\cite{lu2022index} except for the matting task where the `M2O-DIN' version is used. Also, FADE~\cite{lu2022fade} without the gating mechanism is adopted, which we find the performance is more stable across tasks. The hyper-parameter settings in SAPA-D for each model can be referred to Table~\ref{tab:hyper-parameter}. Then, we conduct ablation studies to justify the effectiveness of our design choices. All our experiments are conducted on a server with $8$ NVIDIA GeForce RTX 3090 GPUs. 

\subsection{Semantic Segmentation}
\begin{table}[!t]
    \caption{Semantic segmentation results with SegFormer-B1 on ADE20K. Best performance is in \textbf{boldface} and second best is \underline{underlined}}
    \label{tab:segformer}
    \vspace{5pt}
    \centering
    \renewcommand{\arraystretch}{1.2}
    \addtolength{\tabcolsep}{5pt}
    \begin{tabular}{@{}lllcc@{}}
    \toprule
        SegFormer-B1~\cite{xie2021segformer} & FLOPs & Params & mIoU & bIoU \\
        \midrule
        Bilinear & 15.9 & 13.7 & 42.11 & 28.16 \\
        Deconv & +34.4 & +3.5 & 40.71 & 25.94 \\
        PixelShuffle~\cite{shi2016real} & +34.4 & +14.2 & 41.50 & 26.58 \\
        CARAFE~\cite{jiaqi2019carafe} & +1.5 & +0.4 & 42.82 & 29.84 \\
        IndexNet~\cite{lu2019indices} & +30.7 & +12.6 & 41.50 & 28.27 \\
        A2U~\cite{dai2021learning} & +0.4 & +0.1 & 41.45 & 27.31 \\
        FADE~\cite{lu2022fade} & +2.7 & +0.3 & 43.06 & \underline{31.68} \\
        SAPA-I & +0.8 & +0 & 43.05 & 30.25 \\
        SAPA-B & +1.7 & +0.2 & \underline{43.20} & 30.96 \\
        SAPA-D & +4.4 & +0.2 & \textbf{44.68} & \textbf{31.74} \\
    \bottomrule
    \end{tabular}
\end{table}
\begin{table}[!t]
    \caption{Semantic segmentation results with UPerNet-R50 on ADE20K. Best performance is in \textbf{boldface} and second best is \underline{underlined}}
    \label{tab:upernet}
    \vspace{5pt}
    \centering
    \renewcommand{\arraystretch}{1.2}
    \addtolength{\tabcolsep}{4pt}
    \begin{tabular}{@{}lclcc@{}}
    \toprule
        UPerNet~\cite{zhou2017scene} & Backbone & Params & mIoU & bIoU \\
        \midrule
        Bilinear & R50 & 66.5 & 41.09 & 28.04 \\
        Deconv & R50 & +7.1 & 41.43 & 27.72 \\
        PixelShuffle~\cite{shi2016real} & R50 & +28.3 & 41.35 & 27.49 \\
        CARAFE~\cite{jiaqi2019carafe} & R50 & +0.3 & 41.49 & 28.29 \\
        IndexNet~\cite{lu2019indices} & R50 & +25.2 & 41.42 & 27.88 \\
        A2U~\cite{dai2021learning} & R50 & +0.1 & 41.37 & 27.71 \\
        FADE~\cite{lu2022fade} & R50 & +0.1 & \underline{41.83} & 27.92 \\
        SAPA-I & R50 & +0 & 41.51 & \underline{28.60} \\
        SAPA-B & R50 & +0.1 & 41.47 & 28.27 \\
        SAPA-D & R50 & +0.1 & \textbf{42.60} & \textbf{28.97} \\
        \midrule
        Bilinear & R101 & 85.5 & 43.33 & 30.21 \\
        SAPA-D & R101 & +0.1 & \textbf{44.31} & \textbf{31.47} \\
    \bottomrule
    \end{tabular}
\end{table}


Semantic segmentation 
requires to predict per-pixel class labels such that the pixel group of the same object class is clustered together. 
Generally, the decoder of a segmentation model follows a 
stage-by-stage upsampling architecture, thus the upsampler can play a vital role. This task is rather suitable to justify the upsampling behaviors of SAPA, because one can not only assess the quality of both regional and boundary predictions from numerical metrics but also from qualitative visualizations. 


\subsubsection{Data Sets, Metrics, Baselines, and Protocols}
\label{ssec:dataset}
We conduct experiments on the ADE20K dataset~\cite{zhou2017scene}. 
Apart from the widely-used mIoU metric, we also report the boundary IoU (bIoU)~\cite{cheng2021boundary} metric to evaluate the 
boundary quality.
We consider two baseline models: the Transformer-based model SegFormer-B1~\cite{xie2021segformer} and the CNN-based model UPerNet~\cite{xiao2018unified}.\footnote{In our conference version~\cite{lu2022sapa}, three Transformer-based models, including SegFomer, MaskFormer~\cite{cheng2021per}, and Mask2Former~\cite{cheng2021masked} are considered as baselines. However, later we find that 
the performances of MaskFormer and Mask2Former show large variances, which disturbs the relative importance indication of upsamplers. 
In addition, we 
also overlook CNN-based architectures. As a result, we 
resort to adopt SegFormer and UPerNet as two representative Transformer and CNN baselines, whose performances are relatively stable.
} For SegFormer, we replace the default bilinear interpolation with SAPA in the MLP head. In UPerNet, SAPA is adopted in FPN. Considering the dynamic upsamplers typically use an upscaling factor of $2$, we replace the default evaluation setting with {\tt aligned resize} so that the image size is divisible by $32$. 
We follow the implementation details and the $80K$ iteration training settings in \texttt{mmsegmentation}, and only modify the upsampling stages with specific upsampling operators.

\subsubsection{Semantic Segmentation Results}
Quantitative results are reported in Table~\ref{tab:segformer} and Table~\ref{tab:upernet}. We can observe that SAPA 
consistently improves performance on both SegFormer and UPerNet and outperforms other comparing upsamplers. Even with no parameter, SAPA-I achieves $43.05$ mIoU with SegFormer-B1 and $41.51$ mIoU on UPerNet-R50, which is on par with the previous best upsampler FADE. SAPA-D further pushes the state of the art, which invites $+2.57$ mIoU and $+3.58$ bIoU on SegFormer-B1 and $+1.51$ mIoU and $+0.93$ bIoU on UPerNet-R50. With a stronger backbone R101, SAPA-D on UPerNet also brings around $+1$ mIoU and bIoU improvements. Such results support that SAPA encourages not only regional smoothness but also boundary sharpness. The qualitative results shown in Fig.~\ref{fig:task_visual} also confirm our claim. Moreover, UPerNet adopts more complex structures in the decoder, e.g., PPM~\cite{zhao2017pyramid} and FPN~\cite{lin2017feature}, while SegFormer only use a simple MLP decoder. As a result, the effect of upsamplers in UPerNet is not so obvious as in SegFormer. In addition, we notice that the matting-orientated A2U is the worst performing upsampler on this task, even falling behind bilinear interpolation, which implies some dynamic upsamplers still have a certain task bias.

\subsection{Object Detection}

\begin{table}[!t]
\caption{Object detection results of Faster R-CNN with ResNet-50 on MS-COCO. Best performance is in \textbf{boldface} and second best is \underline{underlined}}
\centering
\renewcommand{\arraystretch}{1.2}
\addtolength{\tabcolsep}{-4.5pt}
\begin{tabular}{@{}lclcccccc@{}}
\toprule
Faster R-CNN & Backbone & Params & $AP$ & $AP_{50}$ & $AP_{75}$ & $AP_S$  & $AP_M$  & $AP_{L}$  \\
\midrule 
Nearest & R50 & 46.8 & 37.4 & 58.1 & 40.4 & 21.2 & 41.0 & 48.1 \\
Deconv & R50 & +2.4 & 37.3 & 57.8 & 40.3 & 21.3 & 41.1 & 48.0 \\
PixelShuffle~\cite{shi2016real} & R50 & +9.4 & 37.5 & 58.5 & 40.4 & 21.5 & 41.5 & 48.3 \\
CARAFE~\cite{jiaqi2019carafe} & R50 & +0.3 & \underline{38.6} & \underline{59.9} & \underline{42.2} & \underline{23.3} & \underline{42.2} & \underline{49.7} \\
IndexNet~\cite{lu2019indices} & R50 & +8.4 & 37.6 & 58.4 & 40.9 & 21.5 & 41.3 & 49.2 \\
A2U~\cite{dai2021learning} & R50 & +0.1 & 37.3 & 58.7 & 40.0 & 21.7 & 41.1 & 48.5 \\
FADE~\cite{lu2022fade} & R50 & +0.2 & 38.5 & 59.6 & 41.8 & 23.1 & \underline{42.2} & 49.3 \\
SAPA-I & R50 & +0 & 37.7 & 59.2 & 40.6 & 22.2 & 41.2 & 48.4\\
SAPA-B & R50 & +0.1 & 37.8 & 59.2 & 40.6 & 22.4 & 41.4 & 49.1 \\
SAPA-D & R50 & +0.6 & \textbf{39.2} & \textbf{60.8} & \textbf{42.7} & \textbf{23.5} & \textbf{42.9} & \textbf{50.3} \\
\midrule
Nearest & R101 & 65.8 & 39.4 & 60.1 & 43.1 & 22.4 & 43.7 & 51.1 \\
SAPA-D & R101 & +0.6 & \textbf{40.6} & \textbf{61.8} & \textbf{44.1} & \textbf{24.3} & \textbf{45.0} & \textbf{52.8} \\
\bottomrule
\end{tabular}
\label{tab:object_detection}
\end{table}

Object detection simultaneously addresses the \textit{where-and-what} problem of objects where each object is labelled with a boundary box and a class label. This task favors both accurate localization and classification. 
Since many existing object detectors employ FPN-like architectures, 
upsamplers can be important 
in acquiring semantically clear feature maps for better localization and classification.

\subsubsection{Data Sets, Metrics, Baselines, and Protocols}
For object detection, we use the MS COCO~\cite{lin2014microsoft} dataset, which involves $80$ object categories. We use AP as the evaluation metric. Among existing 
detectors, we choose the widely-used Faster R-CNN~\cite{ren2015faster} with ResNet-50~\cite{he2016deep} and ResNet-101 as our baseline. 
Over the years, Faster R-CNN has undergone many design iterations. Its performance is stable and has been significantly improved since its original version. We hence validate SAPA based on Faster R-CNN. We use the implementation provided by \texttt{mmdetection}~\cite{mmdetection} and follow its $1\times$ ($12$ epochs) training configurations. We only modify the upsampling stages in FPN.

\subsubsection{Object Detection Results}
Quantitative results are shown in Table~\ref{tab:object_detection}, and qualitative results are visualized in Fig.~\ref{fig:task_visual}. 
SAPA-I and SAPA-B fall behind the former strongest upsampler CARAFE by a large margin. With dynamic point selection, SAPA-D significantly outperforms CARAFE, with $+0.6$ AP metric ($+1.8$ AP than the default upsampler). With R101 as the backbone, SAPA-D consistently improve the AP metric from $39.4$ to $40.6$.

\subsection{Instance Segmentation}

\begin{table}[!t]
\caption{Instance segmentation results of Mask R-CNN with ResNet50 on MS-COCO. Upsampling operators are replaced in FPN. The parameter increment is identical as in Faster R-CNN. Best performance is in \textbf{boldface} and second best is \underline{underlined}}
\centering
\renewcommand{\arraystretch}{1.2}
\addtolength{\tabcolsep}{-4pt}
\begin{tabular}{@{}lcccccccc@{}}
\toprule
Mask R-CNN  & Task & Backbone & $AP$    & $AP_{50}$ & $AP_{75}$ & $AP_S$  & $AP_M$  & $AP_{L}$  \\
\midrule 
Nearest  & Bbox & R50 & 38.3 & 58.7 & 42.0 & 21.9 & 41.8 & 50.2 \\
Deconv  &  & R50 & 37.9 & 58.5 & 41.0 & 22.0 & 41.6 & 49.0 \\
PixelShuffle~\cite{shi2016real}  &  & R50 & 38.5 & 59.4 & 41.9 & 22.0 & 42.3 & 49.8 \\
CARAFE~\cite{jiaqi2019carafe}   & & R50 & \underline{39.2} & 60.0 & \underline{43.0} & 23.0 & \underline{42.8} & 50.8 \\
IndexNet~\cite{lu2019indices} & & R50 & 38.4 & 59.2 & 41.7 & 22.1 & 41.7 & 50.3 \\
A2U~\cite{dai2021learning}      & & R50 & 38.2 & 59.2 & 41.4 & 22.3 & 41.7 & 49.6 \\
FADE~\cite{lu2022fade} & & R50 & 39.1 & \underline{60.3} & 42.4 & \underline{23.6} & 42.3 & \underline{51.0} \\
SAPA-I & & R50 & 38.6 & 59.5 & 42.1 & 23.3 & 42.0 & 50.0 \\
SAPA-B & & R50 & 38.7 & 59.7 & 42.2 & 23.1 & 41.8 & 49.9\\
SAPA-D &  & R50 & \textbf{40.0} & \textbf{61.0} & \textbf{43.5} & \textbf{24.5} & \textbf{43.2} & \textbf{51.8} \\
\midrule
Nearest  &  & R101 & 40.0 & 60.4 & 43.7 & 22.8 & 43.7 & 52.0 \\
SAPA-D &  & R101 & \textbf{41.1} & \textbf{62.3} & \textbf{44.7} & \textbf{24.4} & \textbf{45.2} & \textbf{53.6} \\
\midrule 
Nearest & Segm & R50 & 34.7 & 55.8 & 37.2 & 16.1 & 37.3 & 50.8 \\
Deconv  &  & R50 & 34.5 & 55.5 & 36.8 & 16.4 & 37.0 & 49.5 \\
PixelShuffle~\cite{shi2016real} &  & R50 & 34.8 & 56.0 & 37.3 & 16.3 & 37.5 & 50.4 \\
CARAFE~\cite{jiaqi2019carafe} & & R50 & \underline{35.4} & \underline{56.7} & \underline{37.6} & 16.9 & \underline{38.1} & 51.3 \\
IndexNet~\cite{lu2019indices} & & R50 & 34.7 & 55.9 & 37.1 & 16.0 & 37.0 & 51.1 \\
A2U~\cite{dai2021learning} & & R50 & 34.6 & 56.0 & 36.8 & 16.1 & 37.4 & 50.3 \\
FADE~\cite{lu2022fade} & & R50 & 35.1 & \underline{56.7} & 37.2 & 16.7 & 37.5 & \underline{51.4} \\
SAPA-I & & R50 & 34.9 & 56.4 & 37.4 & \underline{17.1} & 37.7 & 50.3\\
SAPA-B & & R50 & 35.1 & 56.5 & 37.4 & 16.7 & 37.6 & 50.6\\
SAPA-D &  & R50 & \textbf{36.0} & \textbf{58.0} & \textbf{38.1} & \textbf{17.7} & \textbf{38.4} & \textbf{52.1}\\
\midrule
Nearest &  & R101 & 36.0 & 57.6 & 38.5 & 16.5 & 39.3 & 52.2 \\
SAPA-D &  & R101 & \textbf{37.0} & \textbf{59.4} & \textbf{39.3} & \textbf{17.7} & \textbf{40.4} & \textbf{54.0}\\
\bottomrule
\end{tabular}
\label{tab:instance_segmentation}
\end{table}

Instance segmentation is the task of detecting and delineating each distinct object of interest from an image.
Being a unified task of detection and segmentation, 
it has the strong need of both semantic preservation and boundary delineation. Feature upsampling thus matters.

\subsubsection{Data Sets, Metrics, Baselines, and Protocols}
Similar to object detection, we also use the MS COCO~\cite{lin2014microsoft} dataset for instance segmentation. We report the standard Box AP and Mask AP as the evaluation metrics. Mask R-CNN~\cite{he2017mask} with ResNet-50~\cite{he2016deep} and ResNet-101 is adopted as the baseline. Akin to Faster R-CNN, we only modify the upsampling stages in FPN. We also use the codebase of \texttt{mmdetection}~\cite{mmdetection} and follow the default $1\times$ ($12$ epochs) training configurations.

\subsubsection{Instance Segmentation Results}
\begin{figure*}[!t]
	\centering
	\includegraphics[width=\linewidth]{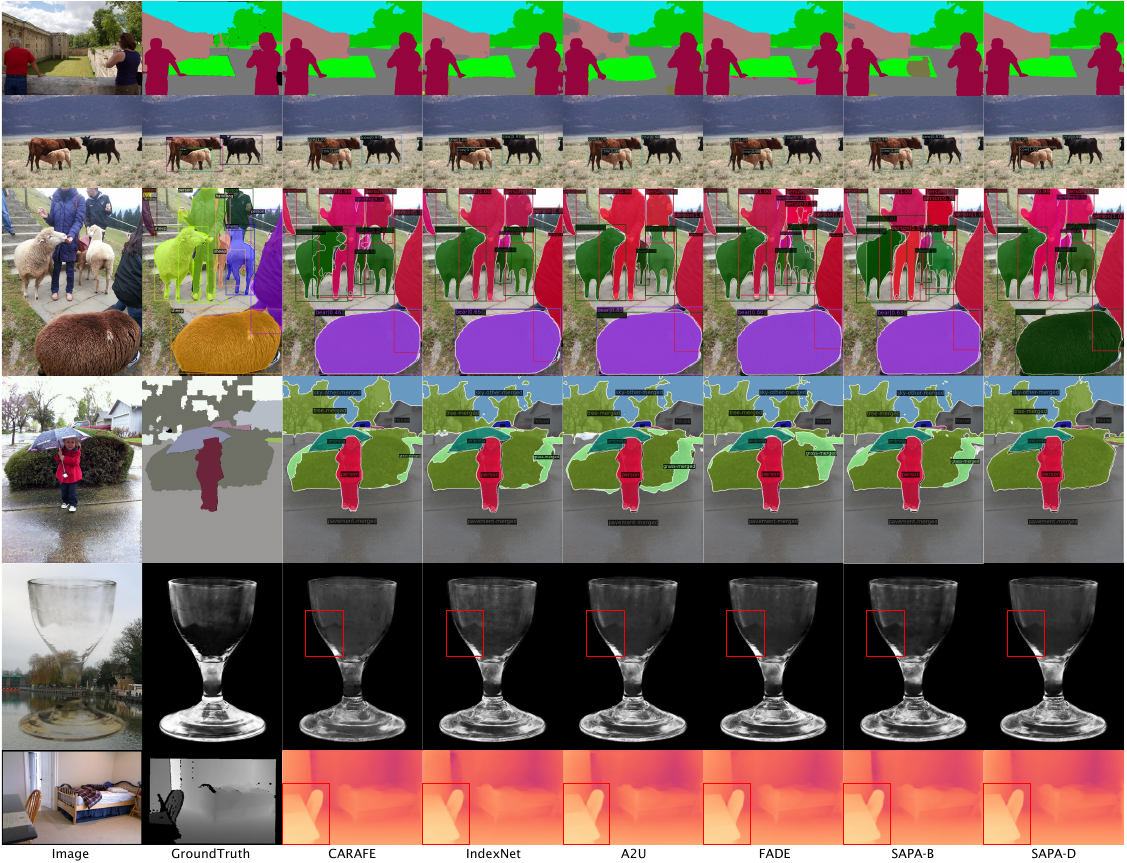}
	\caption{\textbf{Qualitative visualizations on six dense prediction tasks of SAPA and other upsamplers.} The outputs from top to bottom are semantic segmentation, object detection, instance segmentation, ponaptic segmentation, image matting and depth estimation. For semantic segmentation, the masks are outputted by SegFormer-B1. SAPA-B provides shaper boundaries, but hurts the integrity of interior regions. SAPA-D also compensates details, while preserving the continuous interior regions.
	}
	\label{fig:task_visual}
\end{figure*}

Quantitative and qualitative results are shown in Table~\ref{tab:instance_segmentation} and Fig.~\ref{fig:task_visual}, respectively. A similar tendency can be observed as in object detection: all SAPA variants invite consistent performance improvements, and SAPA-D is the best performing variant. In particular, SAPA-D outperforms the previous best baseline CARAFE by $0.6$ Box AP and $0.6$ Mask AP, and introduce a total improvement of $1.7$ Box AP and $1.3$ Mask AP for Mask R-CNN-R50. A similar improvement ($+1.1$ Box AP and $+1.0$ Mask AP) can be observed when ResNet-101 is applied.

\subsection{Panoptic Segmentation}

\begin{table}[!t]
\caption{Panoptic segmentation results with Panoptic FPN on MS-COCO. Upsampling operators are replaced in FPN. Best performance is in \textbf{boldface} and second best is \underline{underlined}}
\centering
\renewcommand{\arraystretch}{1.2}
\addtolength{\tabcolsep}{-3pt}
\begin{tabular}{@{}lclccccc@{}}
\toprule
PanopticFPN & Backbone & Params & $PQ$ & $PQ^{th}$ & $PQ^{st}$ & $SQ$ & $RQ$ \\
\midrule 
Nearest & R50 & 46.0 & 40.2 & 47.8 & 28.9 & 77.8 & 49.3 \\
Deconv & R50 & +1.8 & 39.6 & 47.0 & 28.4 & 77.1 & 48.5 \\
PixelShuffle~\cite{shi2016real} & R50 & +7.1 & 40.0 & 47.4 & 28.8 & 77.1 & 49.1 \\
CARAFE~\cite{jiaqi2019carafe} & R50 & +0.2 & 40.8 & 47.7 & \underline{30.4} & \underline{78.2} & 50.0 \\
IndexNet~\cite{lu2019indices} & R50 & +6.3 & 40.2 & 47.6 & 28.9 & 77.1 & 49.3 \\
A2U~\cite{dai2021learning} & R50 & +0.1 & 40.1 & 47.6 & 28.7 & 77.3 & 48.0 \\
FADE~\cite{lu2022fade} & R50 & +0.1 & 40.9 & 48.0 & 30.3 & 78.1 & 50.1 \\
SAPA-I & R50 & +0 & 40.6 & 47.6 & 29.9 & \underline{78.2} & 49.6 \\
SAPA-B & R50 & +0.1 & \underline{41.0} & \underline{48.1} & 30.2 & 78.1 & \underline{50.2} \\
SAPA-D & R50 & +0.4 & \textbf{42.1} & \textbf{48.4} & \textbf{32.6} & \textbf{79.3} & \textbf{51.3}\\
\midrule
Nearest & R101 & 65.0 & 42.2 & 50.1 & 30.3 & 78.3 & 51.4 \\
SAPA-D & R101 & +0.4 & \textbf{43.5} & \textbf{50.2} & \textbf{33.6} & \textbf{79.2} & \textbf{52.9}\\
\bottomrule
\end{tabular}
\label{tab:panoptic_segmentation}
\end{table}

Panoptic segmentation 
further unifies semantic segmentation and instance segmentation. It enables a holistic view on classifying both stuff and things at the pixel level. Here we evaluate the effect of different upsamplers on this task. 

\subsubsection{Data Sets, Metrics, Baselines, and Protocols}
We adopt the MS COCO~\cite{lin2014microsoft} dataset for panoptic segmentation as well. Instead we report the task-specific $PQ$, $SQ$, and $RQ$~\cite{kirillov2019panoptic} as our evaluation metrics. For the baseline, we choose Panoptic FPN~\cite{kirillov2019panopticfpn} with ResNet-50 and only modify the upsampling stages in FPN. We also use the \texttt{mmdetection}~\cite{mmdetection} codebase and follow the default $1\times$ ($12$ epochs) training configurations.

\subsubsection{Panoptic Segmentation Results}
Quantitative and qualitative results are shown in Table~\ref{tab:panoptic_segmentation} and Figure~\ref{fig:task_visual}, respectively. We observe that, compared with the NN baseline, SAPA-D invites a significant $+1.9$ improvement on the PQ metric. However, despite with a similar network architecture, different upsamplers do not always behave the same in Panoptic FPN as in Mask R-CNN. In Panoptic FPN, while SAPA-D is still the best performing upsampler, SAPA-B already surpasses CARAFE. We think the reason is that boundary precision shows different importance in different tasks. CARAFE has a semantic mending mechanism, so it tends to generate accurate internal predictions for a region while less accurate boundaries. Instead SAPA preserves semantic smoothness but not necessarily rectifies the semantic clusters. Compared with CARAFE, it would generate more accurate boundaries while less accurate regions. Now in instance segmentation, since there exist many isolated instances, incorrect boundaries affect only different instances but not the background. On the contrary, poor boundaries in semantic segmentation are counted twice in performance metrics, on both objects and background. Hence, compared with segmenting instances alone, segmenting stuff in panoptic segmentation actually increases the importance of boundary prediction, which somehow explains why SAPA-B outperforms CARAFE on this task. With a stronger backbone R101, SAPA-D also boosts $+1.3$ PQ metric. Remarkably, on Faster R-CNN, Mask R-CNN and Panoptic FPN, SAPA-D plays a similar role for the performance as replacing the ResNet-50 backbone with ResNet-101.

\subsection{Image Matting}

\begin{table}[!t]
    \caption{Image matting results on the Adobe Composition-1k data set. Best performance is in boldface and second best is \underline{underlined}}
    \centering
    \renewcommand{\arraystretch}{1.2}
    \addtolength{\tabcolsep}{3pt}
    \begin{tabular}{@{}llcccc@{}}
    \toprule
        A2U Matting & Params & SAD & MSE & Grad & Conn\\
        \midrule
        Bilinear & 8.1 & 34.22 & 0.0090 & 17.50 & 31.55 \\
        Deconv & +0.8 & 38.08 & 0.0098 & 19.86 & 36.02 \\
        PixelShuffle~\cite{shi2016real} & +3.2 & \underline{31.15}& 0.0076 & 15.40 & 29.63 \\
        CARAFE~\cite{jiaqi2019carafe} & +0.3 & 32.50 & 0.0086 & 16.36 & 30.35 \\
        IndexNet~\cite{lu2019indices} & +12.3 & 33.36 & 0.0086 & 16.17 & 30.62 \\
        A2U~\cite{dai2021learning} & +0.1 & 32.05 & 0.0081 & 15.49 & 29.21 \\
        FADE~\cite{lu2022fade} & +0.1 & 31.78 & \underline{0.0075} & \underline{15.12} & 28.95 \\
        SAPA-I & +0 & 34.25 & 0.0091 & 18.93 & 32.09 \\
        SAPA-B & +0.1 & 31.19 & 0.0079 & 15.48 & \underline{28.30} \\
        SAPA-D & +0.1 & \textbf{30.68} & \textbf{0.0070} & \textbf{14.12} & \textbf{27.50} \\
    \bottomrule
    \end{tabular}
    \label{tab:matting}
\end{table}

\begin{table*}[!t]
    \caption{Monocular depth estimation results on NYU Depth V2 with DepthFormer. Best performance is in \textbf{boldface} and second best is \underline{underlined}}
    \centering
    \renewcommand{\arraystretch}{1.2}
    \addtolength{\tabcolsep}{0pt}
    \begin{tabular}{@{}l c ccc ccccc@{}}
    \toprule
        DepthFormer-SwinT & Params & $\delta < 1.25$$\uparrow$ & $\delta < 1.25^2$$\uparrow$ & $\delta < 1.25^3$$\uparrow$ & Abs Rel$\downarrow$ & Sq Rel$\downarrow$ & RMS$\downarrow$ & RMS (log)$\downarrow$ & log10$\downarrow$\\
        \midrule
        Bilinear & 47.6 & 0.873 & 0.978 & \underline{0.994} & 0.120 & 0.071 & 0.402 & 0.148 & 0.050\\
        Deconv & +7.1 & 0.872 & \textbf{0.980} & \textbf{0.995} & \underline{0.117} & \textbf{0.067} & 0.401 & 0.147 & 0.050\\
        PixelShuffle~\cite{shi2016real} & +28.2 & 0.874 & \underline{0.979} & \textbf{0.995} & \underline{0.117} & \underline{0.068} & \underline{0.395} & \underline{0.146} & \underline{0.049}\\
        CARAFE~\cite{jiaqi2019carafe} & +0.3 & \underline{0.877} & 0.978 & \textbf{0.995} & \textbf{0.116} & 0.069 & 0.397 & \underline{0.146} & \underline{0.049}\\
        IndexNet~\cite{lu2019indices} & +6.3 & 0.873 & \textbf{0.980} & \textbf{0.995} & \underline{0.117} & \textbf{0.067} & 0.401 & 0.147 & \underline{0.049}\\
        A2U~\cite{dai2021learning} & +0.1 & 0.874 & \underline{0.979} & \textbf{0.995} & 0.118 & \underline{0.068} & 0.397 & 0.147 & \underline{0.049}\\
        FADE~\cite{lu2022fade} & +0.2 & 0.874 & 0.978 & \underline{0.994} & 0.118 & 0.071 & 0.399 & 0.147 & \underline{0.049} \\
        SAPA-B & +0.1 & 0.870 & 0.978 & \textbf{0.995} & \underline{0.117} & 0.069 & 0.406 & 0.149 & 0.050\\
        SAPA-D & +0.7 & \textbf{0.880} & \textbf{0.980} & \textbf{0.995} & \textbf{0.116} & 0.069 & \textbf{0.392} & \textbf{0.144} & \textbf{0.048} \\
    \bottomrule
    \end{tabular}
    \label{tab:depth_estimation}
\end{table*}

In image matting, the model is asked to separate the foreground $F$ from the background $B$ with the alpha matte $\alpha$ from an image $I$ to satisfy the matting equation $I=\alpha F+(1-\alpha)B$. Since alpha mattes often include many subtle details such as hair, furry, and transparent glass, an effective upsampler is required to recover these details from low-res feature maps. Indeed two upsamplers are designed specifically for this task~\cite{lu2019indices,dai2021learning}. The performance on this task can be a good indication on the ability of detail delineation of an upsampler.

\subsubsection{Data Sets, Metrics, Baselines, and Protocols}
We conduct experiments on the Adobe Image Matting dataset~\cite{xu2017deep} and report performance on the Composition-1K testing set. 
The training set consists of $431$ unique foregrounds, and the testing set has $50$ foregrounds and is generated by compositing each foreground with $20$ different backgrounds. 
We report four standard metrics, including Sum of Absolute Difference (SAD), Mean Square Errors (MSE), Gradient errors (Grad), and Connectivity errors (Conn)~\cite{rhemann2009perceptually}. 
The A2U matting~\cite{dai2021learning} is adopted as the baseline. We replace the original upsampler A2U with SAPA in the decoder. In addition, different from~\cite{dai2021learning,lu2022fade}, we empirically find that max pooling is beneficial for image matting, so we replace all the downsampling stages with max pooling (stride=2). We use the code provided by the authors and follow the same training settings as in the original paper.

\subsubsection{Image Matting Results}
Quantitative results are shown in Table~\ref{tab:matting}, and qualitative results are shown in Fig.~\ref{fig:task_visual}. Compared with the strong baseline A2U, SAPA-D reduces the SAD metric by $1.47$ and the Grad metric by $1.82$. We also observe large performance variances in different upsamplers. Being the former strongest upsampler on object detection and instance segmentation, CARAFE conversely is the worst one in image matting, which suggests its sensitivity to details and also its task bias. SAPA-D 
instead exhibits no task bias so far. 

\subsection{Monocular Depth Estimation}

In monocular depth estimation, one needs to obtain the depth of every pixel from a single image. 
This task requires both good region and detail delineation, because an image can include not only flat regions with identical depth but also detailed regions with complicate shapes (\textit{e.g.}, tree branches). Therefore an upsampler 
should take 
both 
cases into account. 

\subsubsection{Data Sets, Metrics, Baselines, and Protocols}
For depth estimation, we use the NYU Depth V2 dataset~\cite{silberman2012indoor} and its default train/test split, which contains $120$ K RGB and depth pairs. We follow the official train/test split as previous work, using $249$ scenes for training and $215$ scenes ($654$ images) for testing, respectively. 
We choose DepthFormer~\cite{li2022depthformer} with Swin-T as the baseline and follow its original training configurations. Both accuracy and error metrics are reported, including the accuracy with threshold $thr$ ($\delta<thr$), root mean squared error (RMS) and its log version (RMS (log)), absolute relative error (Abs Rel), squared relative error (Sq Rel), and average log$_{10}$ error (log10). 
We only modify upsampling stages. Note that here SAPA-I is not applicable due to the mismatched channel number between encoder and decoder features.

\subsubsection{Depth Estimation Results}
Quantitative results are shown in Table~\ref{tab:depth_estimation}. Among all upsamplers, SAPA-D still works the best.  
Fig.~\ref{fig:task_visual} shows that 
depth values 
are versatile, with not only flat and detailed regions but also gradually changing regions (due to perspective changes). Yet, SAPA shows advantages over other upsamplers on, for instance, the consistency of the ceiling, the shape outlines of chairs, and so on.

\subsection{Ablation Study}

We conduct ablation studies to justify each design choice of SAPA. First, we use SAPA-B to evaluate the impact of different normalization functions, the effect of different embedding dimensions in the linear embedding, and the sensitivity of the kernel size of the fixed window kernel. Then, we use SAPA-D to study the effect of the high-res guidance, the DOF of point selection, the initial shape of point selection, the number of selected points under dynamic kernel shapes and the number of groups. In this ablation study, the default settings include SAPA-B with the embedding dimension $d=32$ and the kernel size $K=5$ and SAPA-D with the offset DOF of $s^2$, the number of selected points as $9$ and the number of groups as $4$. We mainly conduct experiments on SegFormer-B1 and Faster R-CNN-R50, whose settings keep the same as in the main experiments. 

\begin{table}[!t]
    \caption{Ablation studies on normalization function with SegFormer-B1.}
    \label{tab:ablation_norm_func}
    \centering
    \renewcommand{\arraystretch}{1.1}
    \addtolength{\tabcolsep}{0pt}
    \begin{tabular}{@{}lccccc@{}}
        \toprule
        norm func & {\rm none} & $e^x$ & {\rm relu} & {\rm sigmoid} & {\rm softplus}\\
        \cmidrule{2-6}
        mIoU & 41.5 & 43.2 & 40.8 & 42.8 & 42.7 \\
        \bottomrule
    \end{tabular}
\end{table}

\begin{table}[!t]
    \caption{Ablation studies on the embedding dimension and kernel size with SegFormer-B1.}
    \label{tab:ablation_dim_size}
    \centering
    \renewcommand{\arraystretch}{1.1}
    \addtolength{\tabcolsep}{-1pt}
    \begin{tabular}{@{}lccclccc@{}}
        \toprule
        embedding dim & 16 & 32 & 64 & kernel size & 3 & 5 & 7\\
        \cmidrule{2-4} \cmidrule{6-8}
        mIoU & 43.0 & 43.2 & 43.2 & mIoU & 43.1 & 43.2 & 42.5 \\
        \bottomrule
    \end{tabular}
\end{table}

\begin{table}[!t]
    \caption{Ablation studies on effect of the high-res guidance, the offset DOF and the ways for offset initialization. The mIoU and AP metric are produced by SegFormer-B1 and Faster R-CNN-R50 respectively.}
    \label{tab:ablation_d}
    \centering
    \renewcommand{\arraystretch}{1.1}
    \addtolength{\tabcolsep}{0pt}
    \begin{tabular}{@{}ccccc@{}}
        \toprule
        guidance & offset DOF & initial & mIoU & AP \\
        \midrule
        \XSolidBrush & $s^2$ & origin & 42.46 & 38.9 \\
        \rowcolor{gray!10}\Checkmark & $s^2$ & origin & 43.11 & \textbf{39.2} \\
        \rowcolor{gray!10}\Checkmark & $1$ & origin & \textbf{44.68} & 38.8 \\
        \Checkmark & $s^2$ & grid & 43.05 & 38.8 \\
        \Checkmark & $1$ & grid & 44.41 & 38.7 \\
        \bottomrule
    \end{tabular}
\end{table}

\begin{table}[!t]
    \caption{Ablation studies on number of points and number of upsampling groups with Faster R-CNN-R50.}
    \label{tab:ablation_point_group}
    \centering
    \renewcommand{\arraystretch}{1.1}
    \addtolength{\tabcolsep}{0pt}
    \begin{tabular}{@{}lcccc@{}}
        \toprule
        Upsampler & \#points & \#groups & AP & latency (ms)\\
        \midrule
        SAPA-D & 9 & 1 & 38.5 & 17.0 \\
        SAPA-D & 9 & 2 & 38.8 & 20.3 \\
        \rowcolor{gray!10}SAPA-D & 9 & 4 & 39.2 & 25.6 \\
        SAPA-D & 9 & 8 & 39.2 & 33.6 \\
        SAPA-D & 2 & 4 & 38.8 & 13.1 \\
        SAPA-D & 4 & 4 & 39.2 & 16.0 \\
        SAPA-D & 16 & 4 & 39.3 & 44.5 \\
        SAPA-D & 25 & 4 & 39.3 & 63.0 \\
        \midrule
        \rowcolor{gray!10}SAPA-B & 25 & 1 & 37.8 & 25.6 \\
        SAPA-B & 25 & 4 & 37.3 & 36.5 \\
        \midrule
        \rowcolor{gray!10}CARAFE & 25 & 1 & 38.6 & 11.5 \\
        CARAFE & 25 & 4 & 38.5 & 16.7 \\
        \bottomrule
    \end{tabular}
\end{table}

\vspace{5pt}
\noindent\textbf{Normalization Function.} We validate different normalization functions in Table~\ref{tab:ablation_norm_func}. `none' indicates no normalization is used. 
Results show that $h(x)=e^x$ is the best choice, which means normalization matters.

\vspace{5pt}
\noindent\textbf{Embedding Dimension and Kernel Size in SAPA-B.} Per Table~\ref{tab:ablation_dim_size}, SAPA-B is not sensitive to the embedding dimension. For the kernel size, if too large kernel size is chosen, the smoothing effect of the kernel increases, which has also a negative influence on sharp boundary.


\vspace{5pt}
\noindent\textbf{Effect of High-res Guidance.} With window-shape kernel, the comparison between CARAFE and SAPA-B may already tell the difference. 
Here we further 
set a baseline for dynamic kernel shape, where dynamic point selection is applied 
to CARAFE 
to compare with SAPA-D. Quantitative results are shown in Table~\ref{tab:ablation_d}, Row 1 vs. Row 2.

\vspace{5pt}
\noindent\textbf{Number of Points and Groups in SAPA-D.} From Table~\ref{tab:ablation_point_group} on can see that multi-group upsampling benefits 
SAPA-D, while introducing little additional latency. With the same $4$ groups, increasing the point number helps little, but brings much more latency. For a better trade-off between performance and speed, we 
recommend $4$ groups and $9$ points as the default setting. Note that the multi-group technique is only beneficial for SAPA-D, and it can even degrade the performance of SAPA-B and CARAFE.

\vspace{5pt}
\noindent\textbf{The Offset DOF in SAPA-D.} We choose a pointwise task--semantic segmentation and a non-pointwise task--object detection to study the effect of the offset DOF. From Table~\ref{tab:ablation_d} Row 2 and Row 3, one can see that DOF=1 works better for semantic segmentation while worse for object detection, and when DOF=$s^2$ the situation reverses. 

\vspace{5pt}
\noindent\textbf{Origin Initialization vs. Grid Initialization of Offsets in SAPA-D.} We study the way for initializing the offsets. Per Row 2 vs. Row 5 and Row 3 vs. Row 4 in Table~\ref{tab:ablation_d}, the origin initialization is better than the grid initialization.

\section{Conclusion}

In this paper, we introduce similarity-aware point affiliation, \textit{i.e.}, SAPA. It not only indicates a lightweight but effective upsampling operator suitable for 
dense prediction tasks, but also expresses a high-level concept that characterizes feature upsampling. We first introduce this concept, which formulates upsampling as designating the corresponding semantic cluster for each upsampled point. Then we indicate that, similar to 
what 
recent kernel-based dynamic upsamplers do, information from a local neighborhood can be used for point affiliation. We then provide our solution using local mutual similarity between encoder and decoder features for upsampling and instantiate an upsampler SAPA. Considering the practical use, we also introduce improvements 
from the view of point sampling. SAPA can serve as a universal substitution for most conventional upsamplers. 
Experiments show that SAPA 
is effective across different tasks and architectures and surpasses or at least is on par with prior upsamplers.

We also note that SAPA adopts the high-res feature map as a guidance for upsampling, which narrows its application scenarios. For example, SAPA cannot be applied to image super-resolution tasks. Moreover, although SAPA has been optimized for the trade-off between performance and complexity, its computational cost is still somewhat expensive. 

\ifCLASSOPTIONcaptionsoff
  \newpage
\fi

\bibliography{egbib}
\bibliographystyle{IEEEtran}
%
%
%
%
%
%
%
%
%
%
%
%
%

%
%
%
%
%
%
%
%
%
%
%



%
%
%



%
%
%
%
%
%
%
%
%

%
%
%
%
%

%

%
%
%

%

%
\end{document}